\pdfoutput=1

\documentclass[11pt]{article}

\usepackage[final]{acl}

\usepackage{times}
\usepackage{latexsym}

\usepackage{hyperref}
\usepackage{url}

\usepackage{amssymb}
\usepackage{graphicx}
\usepackage{subcaption}
\usepackage{tabularx} 
\usepackage{mathptmx}
\usepackage{amsmath, calc}

\usepackage{hyperref}
\usepackage{url}
\usepackage{graphicx}
\usepackage{subcaption}
\usepackage{tabularx}
\usepackage{booktabs}
\usepackage{multirow}
\usepackage{cancel}
\usepackage{booktabs, tabularx} 

\newcommand{\x}{\boldsymbol{x}}
\newcommand{\y}{\boldsymbol{y}}

\usepackage[T1]{fontenc}

\usepackage[utf8]{inputenc}

\usepackage{microtype}

\usepackage{inconsolata}

\usepackage{graphicx}

\title{Investigating Factuality in Long-Form Text Generation: The Roles of Self-Known and Self-Unknown}

\author{
Lifu Tu, \ Rui Meng\thanks{Now at Google.}, \ Shafiq Joty, \ Yingbo Zhou, \ Semih Yavuz  \\
Salesforce AI Research\\
\\}

\begin{document}
\maketitle
\begin{abstract}
Large language models (LLMs) have demonstrated strong capabilities in text understanding and generation. However, they often lack factuality, producing a mixture of true and false information, especially in long-form generation. In this work, we investigates the factuality of long-form text generation across various large language models (LLMs), including GPT-4, Gemini-1.5-Pro, Claude-3-Opus, Llama-3-70B, and Mistral. Our analysis reveals that factuality tend to decline in later sentences of the generated text, accompanied by a rise in the number of unsupported claims. Furthermore, we explore the effectiveness of different evaluation settings to assess whether LLMs can accurately judge the correctness of their own outputs: Self-Known (the percentage of supported atomic claims, decomposed from LLM outputs, that the corresponding LLMs judge as correct) and Self-Unknown (the percentage of unsupported atomic claims that the corresponding LLMs judge as incorrect). 
Empirically, we observe a positive correlation between higher Self-Known scores and improved factuality, whereas higher Self-Unknown scores are associated with reduced factuality. Interestingly, the number of unsupported claims can increase even without significant changes in a model’s self-judgment scores (Self-Known and Self-Unknown), likely as a byproduct of long-form text generation. We also derive a mathematical framework linking Self-Known and Self-Unknown scores to factuality: $\textrm{Factuality}=\frac{1-\textrm{Self-Unknown}}{2-\textrm{Self-Unknown}-\textrm{Self-Known}}$, which aligns with our empirical observations. Additional Retrieval-Augmented Generation (RAG) experiments further highlight the limitations of current LLMs in long-form generation and underscore the need for continued research to improve factuality in long-form text. 
\end{abstract}

\section{Introduction}

The long-context capabilities of large language models (LLMs) ~\citep{openai2023gpt4,llama3modelcard,jiang2024mixtral,geminiteam2024gemini,claude3} have seen significant advancements in recent years. Lots of work~\citep{shaham-etal-2023-zeroscrolls,bai-etal-2024-longbench,an-etal-2024-l,zhang-etal-2024-bench,kuratov2024babilongtestinglimitsllms} have explored the ability of LLMs to handle long contexts, however, relatively few have examined their ability for long-form text generation. 

Despite LLMs have the impressive generative abilities, these models are prone to producing hallucinations~\citep{li-etal-2023-halueval,min-etal-2023-factscore} where the generated content often blends factual and fabricated information. This tendency not only undermines performance but also poses substantial risks in practical applications. To assess the factuality of responses from LLMs, recent research~\citep{fan-etal-2020-generating,wright-etal-2022-generating,min-etal-2023-factscore,manakul-etal-2023-selfcheckgpt} has introduced a method that breaks down generations into atomic claims -- short statements each containing a single piece of information. These atomic claims are then individually evaluated to determine whether they are supported by evidence or unsupported.

To ensure the reliable use of LLMs, it is also crucial that they possess the ability to recognize not only "what they know" but also "what they don't know." Recent studies, such as those by \citet{kadavath2022language,liu-etal-2022-token,guerreiro-etal-2023-looking}, have shown that language models can assess the validity of their own claims. However, \citet{srivastava2023imitation,yin-etal-2023-large} have pointed out the limitations of LLMs in recognizing their own knowledge gaps. 


In this work, we investigate the factuality of long-form text generation across various LLMs. We first check the factuality of long-form generation at different relative positions using two annotated datasets and two models: ChatGPT and PerplexityAI (which integrates a search engine). Our findings verify that sentences generated earlier in the sequence generally demonstrate higher factuality. However, these later-generated sentences contain more unsupported claims and fewer supported claims.

To explain this phenomenon, we try to examine whether LLMs exhibit reduced knowledge in later generations with wo metrics: the \textbf{Self-Known score} (the percentage of supported atomic claims judged as correct by the LLMs) and the \textbf{Self-Unknown score} (the percentage of unsupported atomic claims judged as incorrect by the LLMs). These two metrics are used to quantify the corresponding models’ ability to judge the correctness of atomic claims. In order to accurately compute the two metrics, we have tried three different approaches, one of which is a novel approach that adds an answer option: `None of the above'. This modification appears to provide a more accurate measure of the LLMs' abilities, as evidenced by a higher flip rate for supported claims and an increasing flip rate at later relative positions. This suggests that the model reassesses its confidence when faced with an option signaling uncertainty. In contrast, the low flip rate for unsupported claims indicates a consistent judgment of their incorrectness. These results suggest a nuanced understanding by LLMs of supported versus unsupported claims and underscore the importance of specific evaluation settings to accurately gauge model performance. Our findings align with human annotations for two LLMs, although some discrepancies, particularly with the PerplexityAI model, suggest gaps in estimation.

Later, we apply this modified approach to compute the Self-Known and Self-Unknown scores across various LLMs, including GPT-4,
Gemini-1.5-Pro, Claude-3-Opus, Llama-3-70B,
and Mistral. We also develop a mathematical framework that links these scores to factuality. Overall, both empirical and theoretical results demonstrate a strong relationship between factuality and the Self-Known and Self-Unknown scores. We observe that these scores can vary significantly across different models. However, even when the Self-Known and Self-Unknown scores remain relatively stable, the number of unsupported claims tends to increase in later parts of the generated text. This suggests that lower factuality in later sentences is not solely due to score changes, but also influenced by error propagation and diminished model knowledge over time.


The main contributions of our work are as follows:

1. We explore the factuality patterns of long-form text generation across various model families (GPT-4, Gemini-1.5-Pro, Claude-3-Opus, Llama-3-70B, and Mistral). We find that even the most advanced LLMs typically exhibit lower factuality scores in the later segments of long-form text. Retrieval-Augmented Generation (RAG) systems show a similar trend, although they tend to maintain higher factuality overall.

2. We analyze Self-Known and Self-Unknown ratios across different segments of generated text. While Self-Known scores are relatively high, even the strongest LLMs (e.g., GPT-4, Gemini-1.5-Pro, Claude-3-Opus) average only around 50\%, with Self-Unknown scores remaining well above zero. This suggests that even advanced models struggle to recognize the limits of their own knowledge.

3. We develop a mathematical framework linking Self-Known and Self-Unknown scores to factuality. Empirical and theoretical results show higher Self-Known scores improve factuality, while higher Self-Unknown scores reduce it. Notably, unsupported claims can increase even without major changes in self-judgment, highlighting challenges in long-form generation.

4. We find that Retrieval-Augmented Generation (RAG), which supplies needed knowledge, can improve factuality. However, it fails to fully address the issue of lower factuality at a later position. This highlights the need for alternative framework specifically designed for long-form generation tasks.

\section{Long-Form Text Generation}
\label{sec:2}
To evaluate the factuality of LLM responses, recent work~\citep{liu-etal-2023-revisiting,chen-etal-2022-generating,min-etal-2023-factscore} breaks a generation into a series of atomic claims—short statements that each contain one piece of information. Each atomic claim is then individually evaluated to determine whether it is supported or unsupported. In this section, we first explore the factuality patterns of these atomic claims in long-form text generation.
\subsection{Observations}

In order to explore the factuality of long-form generation at different relative positions, we use the human annotated data from~\citet{min-etal-2023-factscore} to compute the macro-average percentage of three different claims (supported, unsupported, and irrelevant) across five different relative positions. In their human-annotated data, each long LLM generation is decomposed into atomic claims and each atomic claim is assigned with one of the three labels (``supported'', ``not-supported'', ``irrelevant'').

The detailed procedures for computing fractions of different type claims at different relative positions are as following: 
1) Calculate the fraction of supported, unsupported, and irrelevant claims for each sentence;  2) Determine each sentence’s relative position in the generation , e.g., if it is the third sentence out of six, its relative position would be 3/6 = 50\%; 3) Group sentences into relative position ranges: 0-20\%, 20\%-40\%,, etc.; 4) Compute the macro-average claim percentages within each group
 

Figure~\ref{fig:relative_position} (a) shows ChatGPT results (PerplexityAI results are in the Appendix). Unsupported claims increase in later-generated sentences. Figure~\ref{fig:relative_position} (b) further shows that as generation continues, LLMs produce more unreliable and unsupported claims.

\paragraph{Open Questions.} Is the phenomenon above due to LLMs having less knowledge about later generations? Can LLMs recognize when claims are supported and when they are not? Do LLMs identify unsupported claims more effectively when they appear later in the text compared to earlier ones?

\begin{figure*}[h!]
\begin{center}
    
\begin{subfigure}[b]{0.48\linewidth}
    \includegraphics[width=\linewidth]{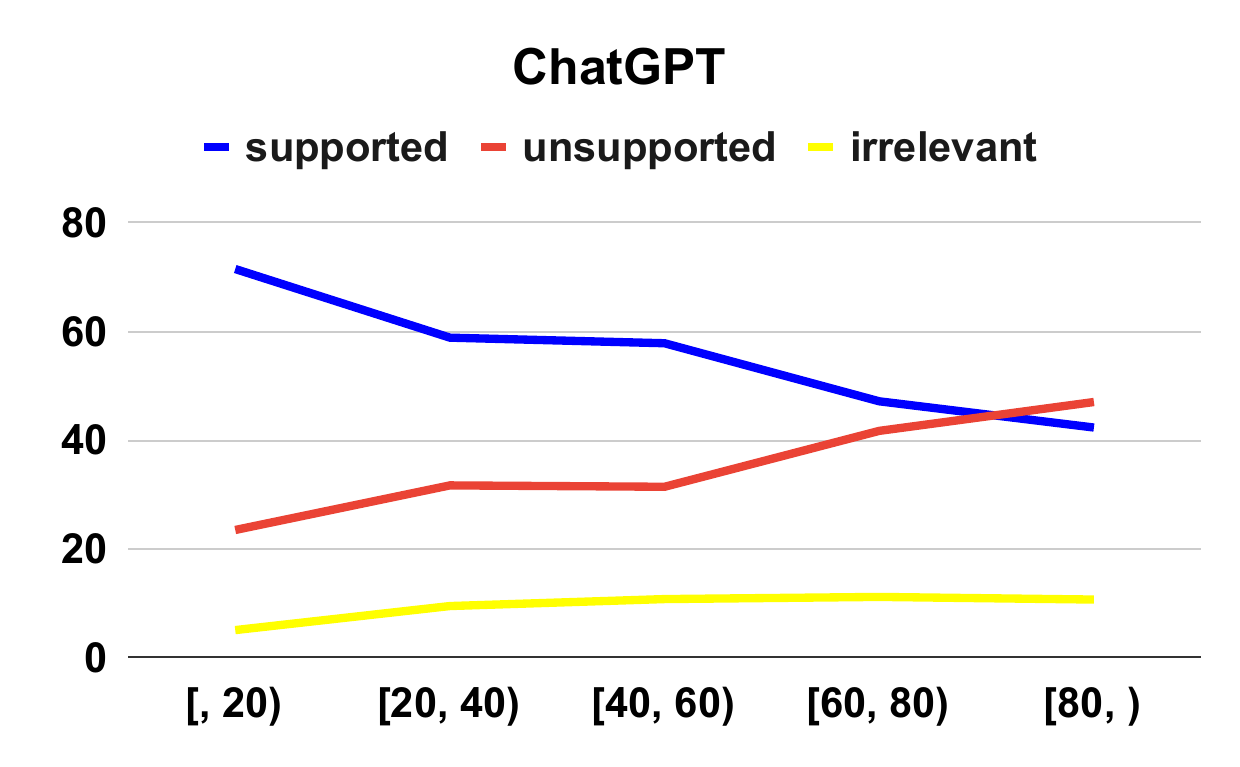}
    \caption{Percentage (\%) of supported, unsupported and irrelevant atomic claims.}
    \label{fig:rNg1}
  \end{subfigure}\hfill
    \begin{subfigure}[b]{0.48\linewidth}
    \includegraphics[width=\linewidth]{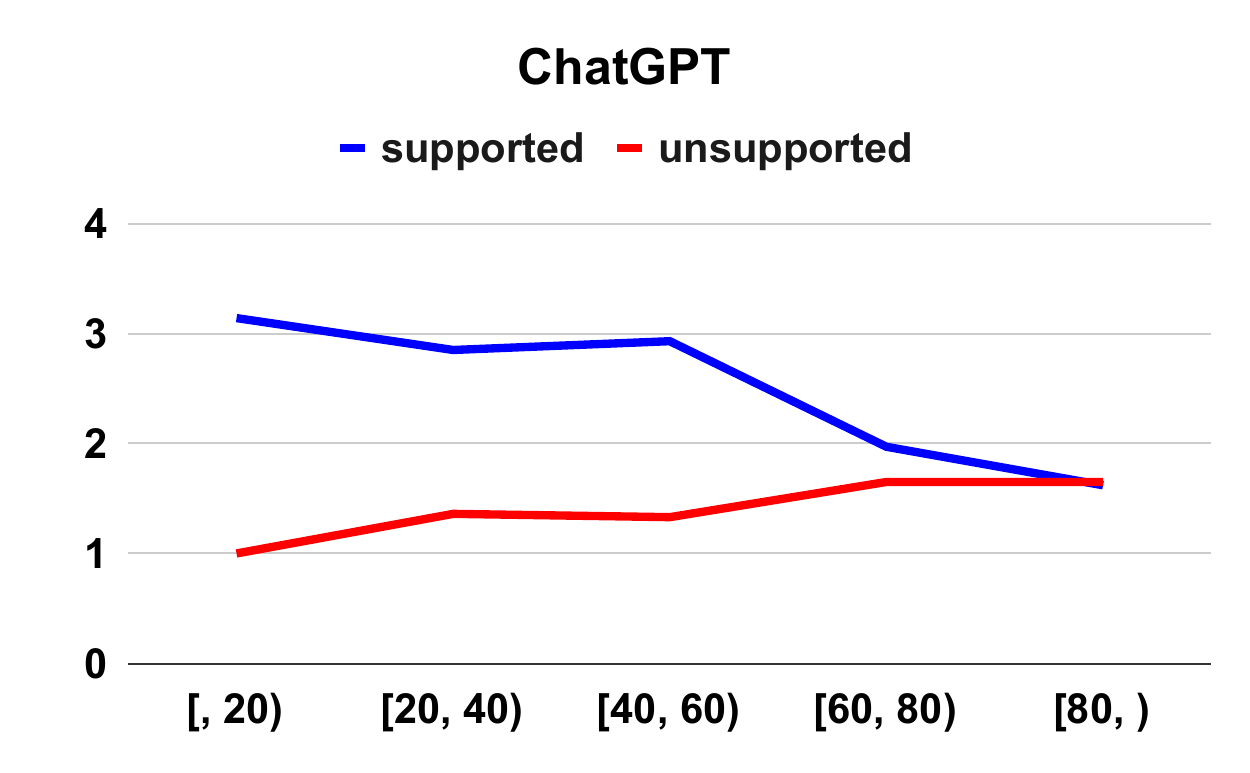}
    \caption{Number of supported and unsupported atomic claims.}
    \label{fig:rNg2}
  \end{subfigure}

\end{center}
\caption[Two numerical solutions]{Long-form generation across different relative positions (\%) for ChatGPT.
}
\label{fig:relative_position}
\end{figure*}







\section{Self-Known and Self-Unknown}
\label{sec:knownUnknown}
To investigate these questions, we examine whether the corresponding LLMs recognize their atomic claims by computing two metrics: \textbf{Self-Known} (the percentage of supported atomic claims that the corresponding LLMs judge as correct) and \textbf{Self-Unknown} (the percentage of unsupported atomic claims that the corresponding LLMs judge as incorrect). While there is related work, such as~\citet{rajpurkar-etal-2018-know,xiong2024can}, our approach differs in two key ways: (1) Evaluation is conducted on atomic claims, which are derived from sentences in long-form generation, rather than assigning a score to the entire model output; (2) Our focus is on factuality (whether an atomic claim is true or false), rather than on uncertainty scores (i.e., "How likely is the above answer to be correct?").

We explore the computation of \textbf{Self-Known} and \textbf{Self-Unknown} using the following three approaches ( with the corresponding prompt templates provided in Appendix Section~\ref{sec:appendix}):
\begin{itemize}
    \item \textbf{Direct-Asking}: In this approach~\citep{rajpurkar-etal-2018-know}, the atomic claim is directly given to the corresponding LLMs and be asked whether the statement is true or false.
    \item \textbf{Question-Answering}: Given an atomic claim, a question-answer pair can be derived~\citep{trischler-etal-2017-newsqa,rajpurkar-etal-2018-know,hu2024towards} with GPT-4 Turbo. For example, "Lanny Flaherty is an American." can be used to derived a question-answer pair ("What nationality is Lanny Flaherty?", "American"). Then, given the question and answer, we ask the corresponding LLMs whether the answer is true or false.
    \item \textbf{Question-Answering w/NOA}: Similar to the above approach, a question-answer pair is derived according to each atomic claim. One big different is: given question and answer, one more addition choice ( "None of the above")~\citep{rajpurkar-etal-2018-know} is given to the corresponding LLMs. This is a well-defined evaluation because it can check whether the model actually knows the answer of the question,  especially if the question is vague or context-information is missing. 
\end{itemize}


\begin{figure*}[h!]
\small
\centering
  \begin{subfigure}[b]{0.49\linewidth}
    \includegraphics[width=\linewidth]{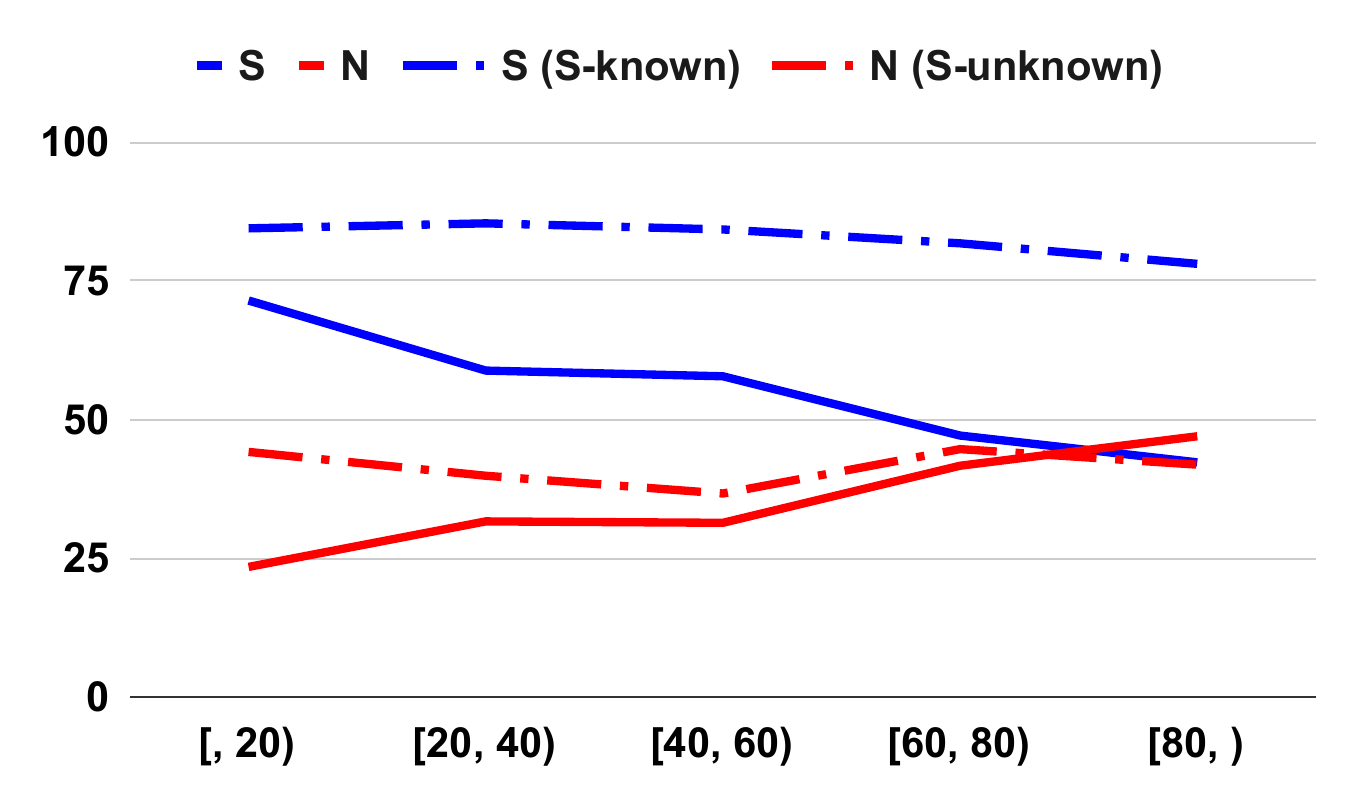}
    \caption{Direct-Asking }
  \end{subfigure}
\begin{subfigure}[b]{0.49\linewidth}
    \includegraphics[width=\linewidth]{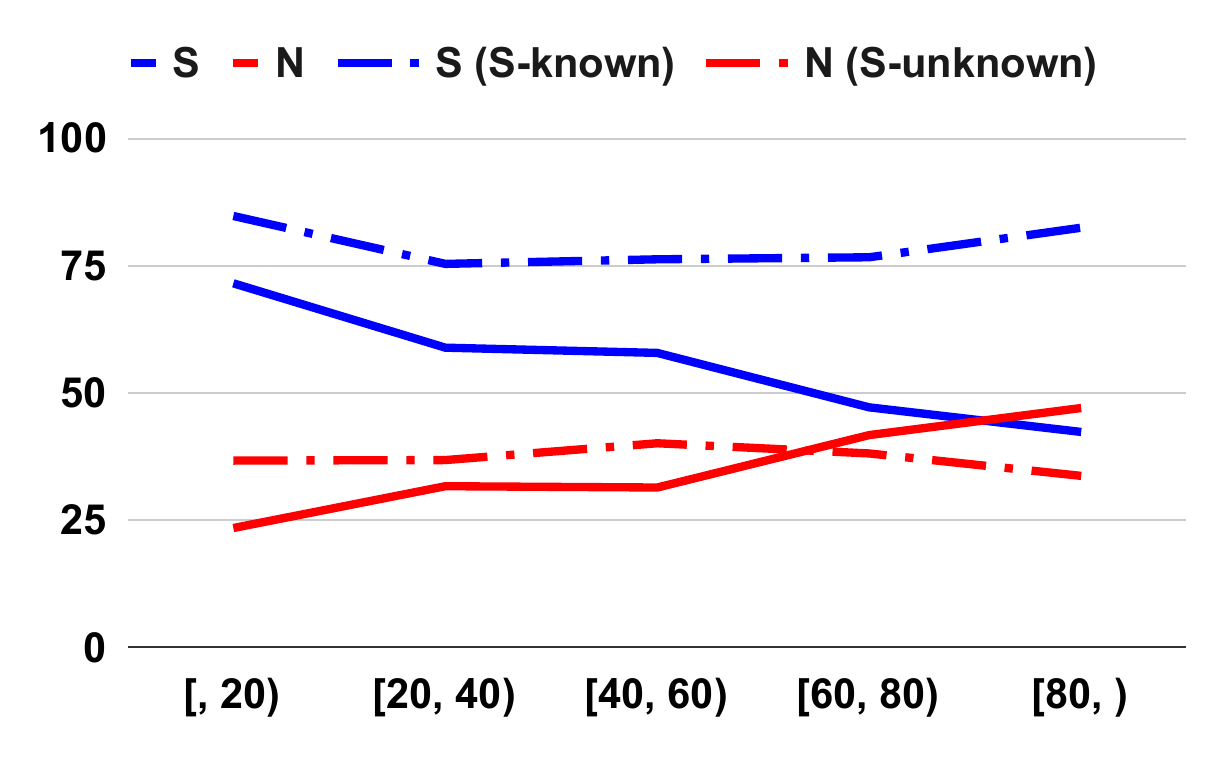}
    \caption{Question-Answering}
  \end{subfigure}
  \begin{subfigure}[b]{0.48\linewidth}
    \includegraphics[width=\linewidth]{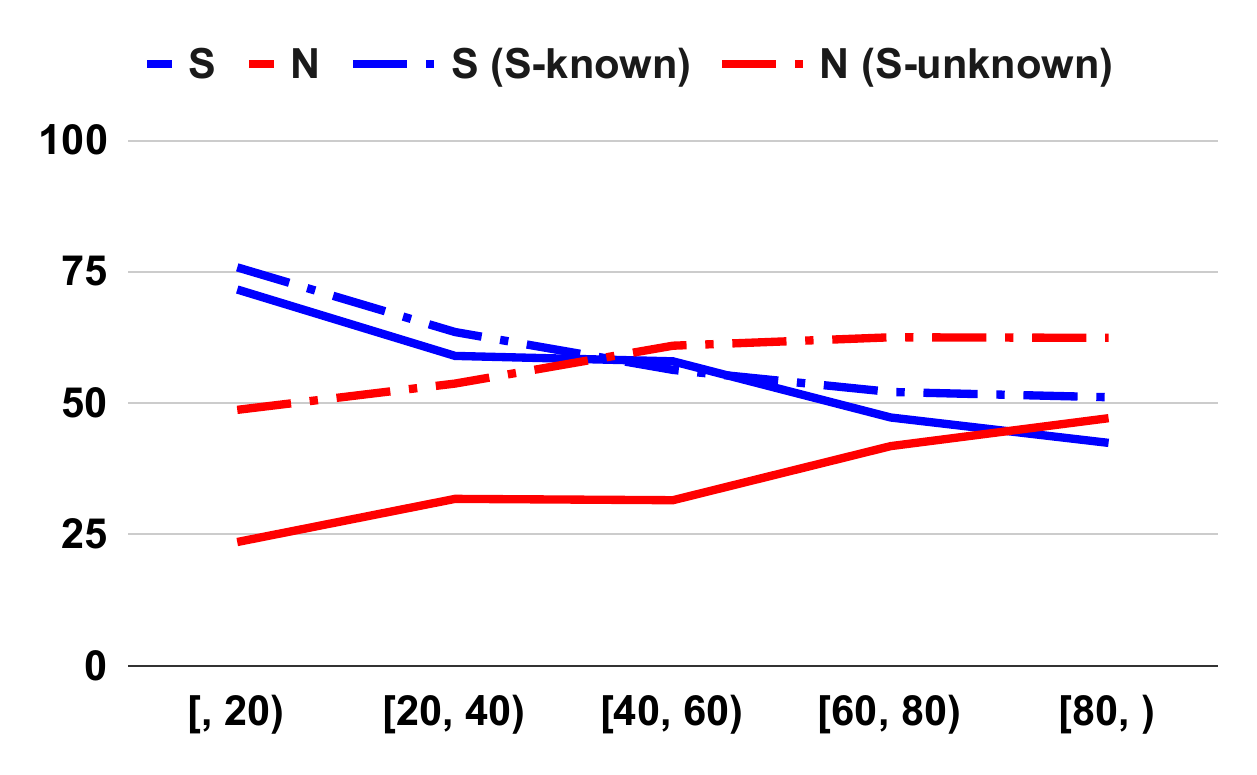}
    \caption{Question-Answering W/ NOA}
  \end{subfigure}
  \begin{subfigure}[b]{0.48\linewidth}
    \includegraphics[width=\linewidth]{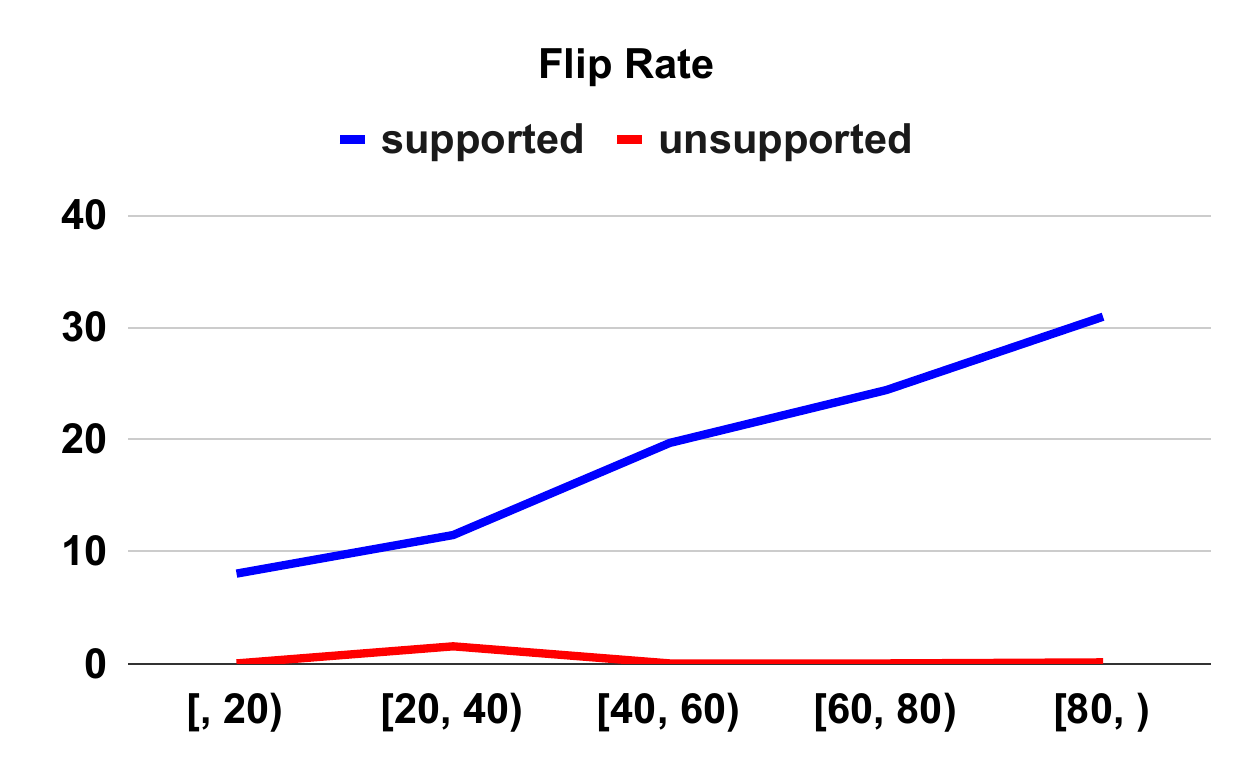}
    \caption{Flip rate (\%) from setting (b) to setting (c) across different relative positions (\%) for both supported claims and unsupported claims. The high flip rate of supported claims indicates that prompting with setting (c) better recognizes whether LLMs accurately assess their knowledge of atomic claims.}
    \label{fig:flipRate}
  \end{subfigure}
  \caption{Self-Know and Self-Unknown results of ChatGPT across different relative positions (\%). \textbf{S}: \textbf{factuality} (percentage of supported atomic claims); \textbf{N}: percentage of unsupported atomic claims; \textbf{S (S-known)}: \textbf{Self-Known} score; \textbf{N (S-unknown)}: \textbf{Self-Unknown} score 
  \label{fig:ChatGPT}
  }
  
\end{figure*}

 We compute the Self-Known score and the Self-Unknown score using these prompt templates. The human annotated data on ChatGPT\footnote{The labeled ChatGPT data is also from ~\citet{min-etal-2023-factscore} as above. There are 183 long generations of ChatGPT.} are used in this experiments. 
 Figure~\ref{fig:ChatGPT} presents the results on ChatGPT. 

\paragraph{Comparison on the above three evaluation settings} With the first two settings, the results of \textbf{Self-Known} score and \textbf{Self-Unknown} score are similar. However, the results of the third setting differ from the other two. We hypothesize that the reason is that the added choice, ``None of the above'' which allows the LLM to determine whether it knows the answer to the question.


To examine the effect of this setting, we plot the flip rate (claims judged as correct by the LLM in setting (b) but judged as incorrect in setting (c)) for supported and unsupported claims. As shown in Figure~\ref{fig:flipRate}, there is a high flip rate for supported claims, and this rate increases with higher relative positions. In contrast, there is almost no flipping for unsupported claims. Therefore, setting (c) is more suitable for checking whether the LLM knows a atomic claim. The high flip rate observed for supported claims suggests that the model is reconsidering its initial judgments when presented with the option ``None of the above''. This indicates that the model may not be entirely confident in its original answers and is more likely to recognize uncertainty. The increasing flip rate for higher relative positions further supports this, implying that the model's confidence decreases as the position of the claim within the context changes.

In summary, we observed similar results between setting (a) (Direct-Asking) and setting (b) (Question-Answering), and a significant difference between setting (b) (Question-Answering) and setting c (Question-Answering W/ NOA). \textbf{The deeper analysis between setting (b) and setting (c) revealed that setting (c) recognizes atomic claims more confidently and treats atomic claims that flip as unknown. This is why we chose to use setting (c) in the subsequent experiments.}




\section{Analysis }
We denote the prompt input of LLMs as $\x$ and long output of LLMs as $\y$. The binary auxiliary label $d=1$ indicates the LLM output is factual correct and $d=0$ indicates LLM output is wrong. 

We assume that $P (d=1 \mid \y, \x)$ is equal to \textbf{factuality score}\footnote{This is an assumption we are making: that there is no overconfidence, and the confidence score is approximately equal to the factuality score.} $\sigma$ of LLM output $\y$. Given $\x$, the joint distribution of between the auxiliary label and model output $(d, \y)$ is
\begin{align}
& \sigma * P(\y \mid \x) \label{eq:know1} \\
=& P(d=1 \mid \y, \x)* P(\y \mid \x) = P(d=1,\y \mid 
\x)  \nonumber \\ 
=& P(d=1, \mathrm{y_{correct}} \mid \x) * \sigma + \nonumber \\ &  P(d=1, \mathrm{y_{wrong}} \mid \x) * (1-\sigma) \nonumber \\
=& P(d=1 \mid \mathrm{y_{correct}} ) * P(\mathrm{y_{correct}} \mid \x) * \sigma +  \nonumber \\ & P(d=1 \mid \mathrm{y_{wrong}} ) * P(\mathrm{y_{wrong}} \mid \x) * (1-\sigma) \label{eq:know}
\end{align}
\noindent $\mathrm{y_{correct}}$ refers to model outputs aligned with the ground truth and $\mathrm{y_{wrong}}$ refers to outputs that are wrong.
Because $\y$ is the generated output according to the log-likelihood, the correct part and incorrect part have similar log-likelihood. Then, it is reasonable to have this following assumption:
\begin{equation}
    P(\y \mid \x) \approx P(\mathrm{y_{correct}} \mid \x) \approx P(\mathrm{y_{wrong}} \mid \x) \nonumber
\end{equation}

Then, after cancel the above three terms in Equation~\ref{eq:know1} and Equation~\ref{eq:know} ,
\begin{align}
    \sigma= & P(d=1 \mid \mathrm{y_{correct}} )  \sigma +  P(d=1 \mid \mathrm{y_{wrong}}) (1-\sigma) \nonumber
\end{align}

\noindent We denote $P(d=1 \mid \mathrm{y_{correct}} )$ and $P(d=0 \mid \mathrm{y_{wrong}} )$ as \textbf{Self-Known} score (percentage of supported atomic claims
judged as correct by LLMs) and \textbf{Self-Unknown} score (percentage of unsupported atomic claims judged as incorrect by LLMs) respectively. Once the above formula is solved, we can determine the relationship among the factuality score, Self-Known score, and Self-Unknown score:
\begin{align}
    \sigma = \frac{1-\textrm{Self-Unknown}}{2-\textrm{Self-Unknown}-\textrm{Self-Known}}\label{eq:factscore}
\end{align}
\noindent Where $\sigma$ is the factuality score.

\paragraph{Factuality Vs. \textbf{Self-Known} Vs. \textbf{Self-Unknown}} Given $\textrm{Self-Unknown} \in [0,1]$ and $\textrm{Self-Known}\in [0,1]$, the factuality score increases when the \textbf{Self-Known} score is increased or the \textbf{Self-Unknown} score is decreased. This matches our observations in Section~\ref{sec:knownUnknown} and Figure~\ref{fig:ChatGPT} (c).

\begin{figure}[h]
\begin{center}
  \includegraphics[width=0.9\columnwidth]{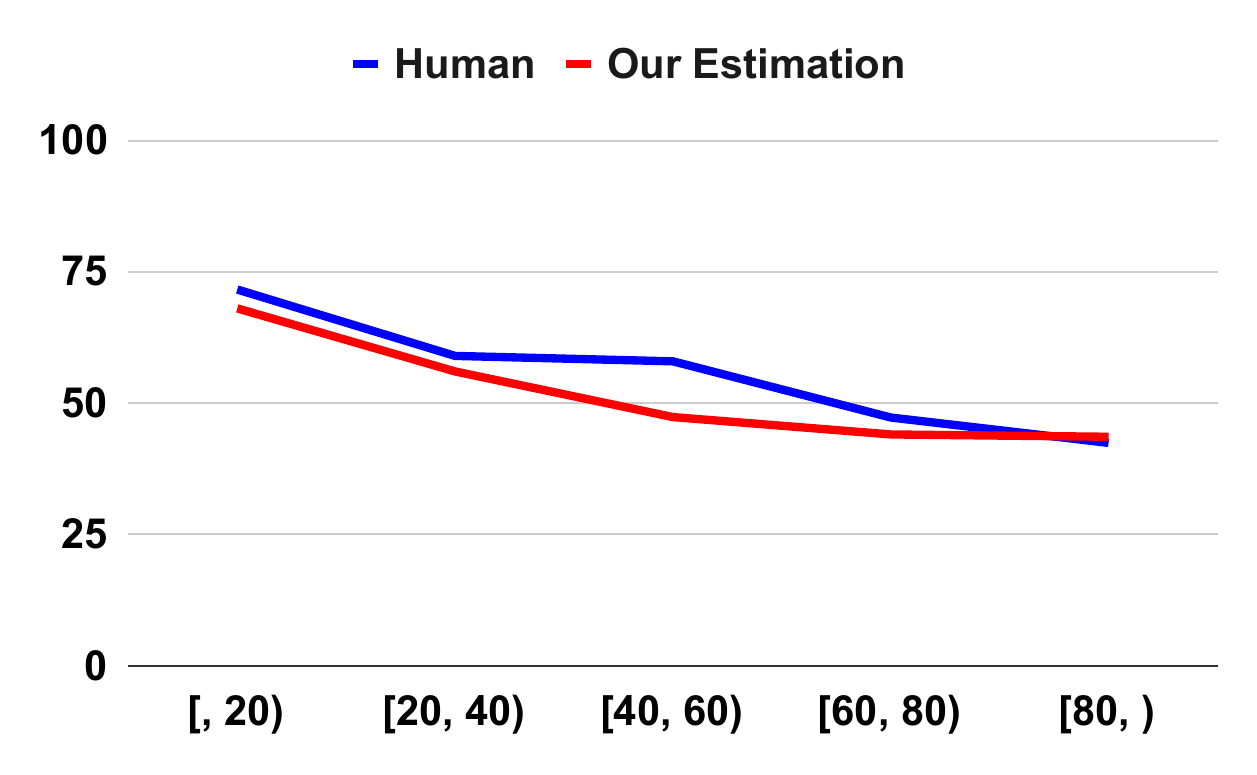}
  \caption{Human-annotation factuality score (\%) and our estimation with Equation~\ref{eq:factscore} across different relative positions (\%).}
  \label{fig:factEstimation}
\end{center}
\end{figure}

\paragraph{Estimation of factuality Score}
In Equation~\ref{eq:factscore}, we present a method for estimating the factuality score. We use the Self-Known and Self-Unknown results of the corresponding model (ChatGPT) with configuration (c) to estimate the factuality score across different relative positions. As shown in Figure~\ref{fig:factEstimation}, our estimation closely matches the human-annotation results.

\section{Automatic Results on Additional LLMs }


In the previous section, our experiments were conducted using human-annotated factuality data. In this section, we first introduce an automated tool for factuality evaluation. Then, using the proposed approach from Section 3 to compute Self-Known and Self-Unknown scores, we analyze the trends in factuality, Self-Known, and Self-Unknown scores across other advanced LLMs

\subsection{Automatic Tool Setting}

In Section~\ref{sec:2}, we used the human annotated data (atomic claims are short statements that are decomposed from the model's generation, and each atomic claim is labeled as either supported or unsupported based on its factual correctness.). 

\paragraph{Configuration} We use the tool FActScore~\citep{min-etal-2023-factscore} for factuality evaluation with the following configuration: the latest version of GPT-3.5 (gpt-3.5-turbo-0125) is used to break a generated text into a series of atomic claims and evaluate each atomic claim against a retrieved knowledge (model name ``retrieval+llama+npm'' is used during the evaluation)\footnote{In the original work, text-davinci-003 was used to get atomic claims and ChatGPT is used to evaluate whether each atomic is supported or unsupported. }.

\paragraph{Results}
 Figure~\ref{fig:tool_relative_position} in the Appendix shows the comparison between the tool's evaluation and human annotation results. We notice the tool's estimation is highly correlate well with human annotations. For number of atomic claims, the absolute difference is not bigger than 1. And the trend of tool's estimation is almost the same as human annotation. For factuality estimation, the tool's results are well-aligned with human annotations for two OpenAI models. Although there is an estimation gap for the PerplexityAI model, the trend of the estimation remains consistent with human annotations.

\paragraph{Takeaway.} The tool with above configurations can well capture the trend of number of atomic claim and factuality.

\subsection{Additional LLMs}

In this section, we explore the factuality of long-form text generation across different relative positions using automatic tools.

\subsubsection{Experimental Setup} 
\label{Exp:setup}
For each LLM, we follow four key steps to obtain experimental results: (1) generating text outputs; (2) filtering the generated content; (3): evaluating factuality; and (4): estimating \textbf{Self-Known} and \textbf{Self-Unknown} scores with the corresponding LLM. For more details on each step, please refer to Appendix Section~\ref{app:steps}.

\begin{figure*}[h!]
\begin{center}
 \small
    \begin{subfigure}[b]{0.45\linewidth}
    \includegraphics[width=\linewidth]{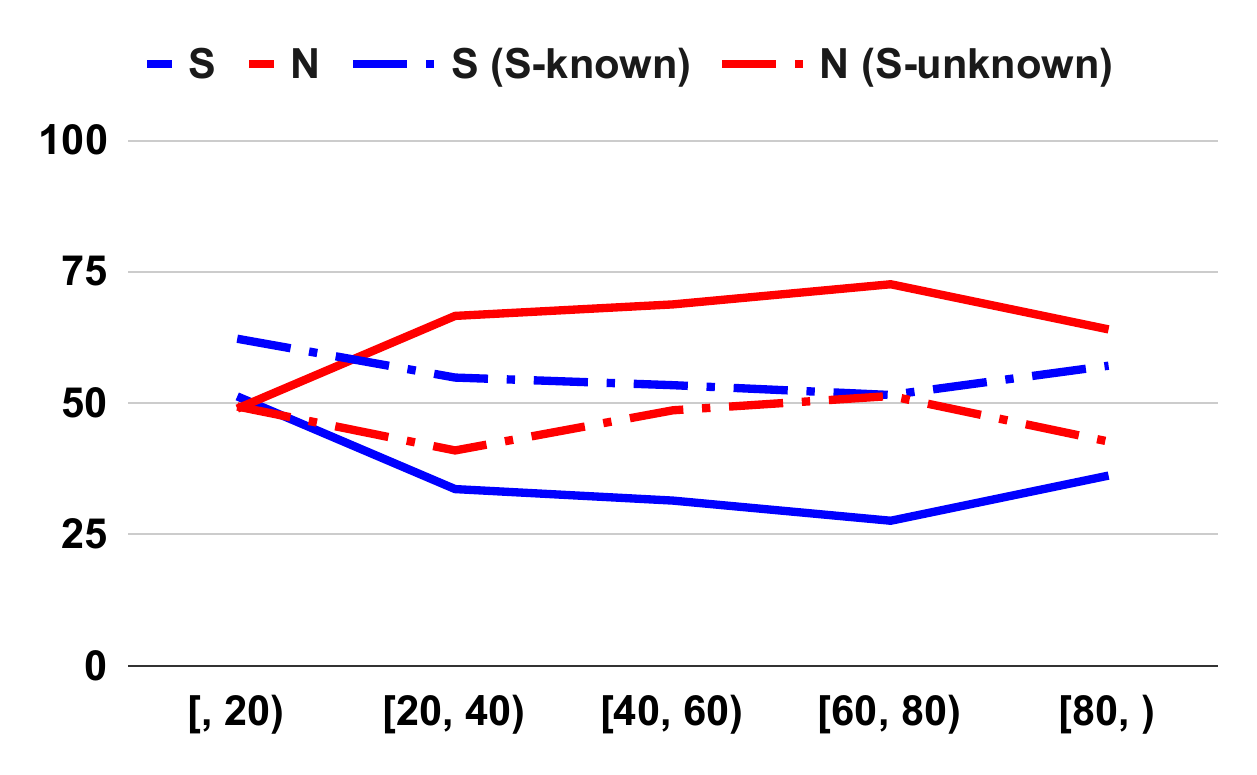}
    \caption{Gemini-1.5-pro}
  \end{subfigure}
  \begin{subfigure}[b]{0.45\linewidth}
    \includegraphics[width=\linewidth]{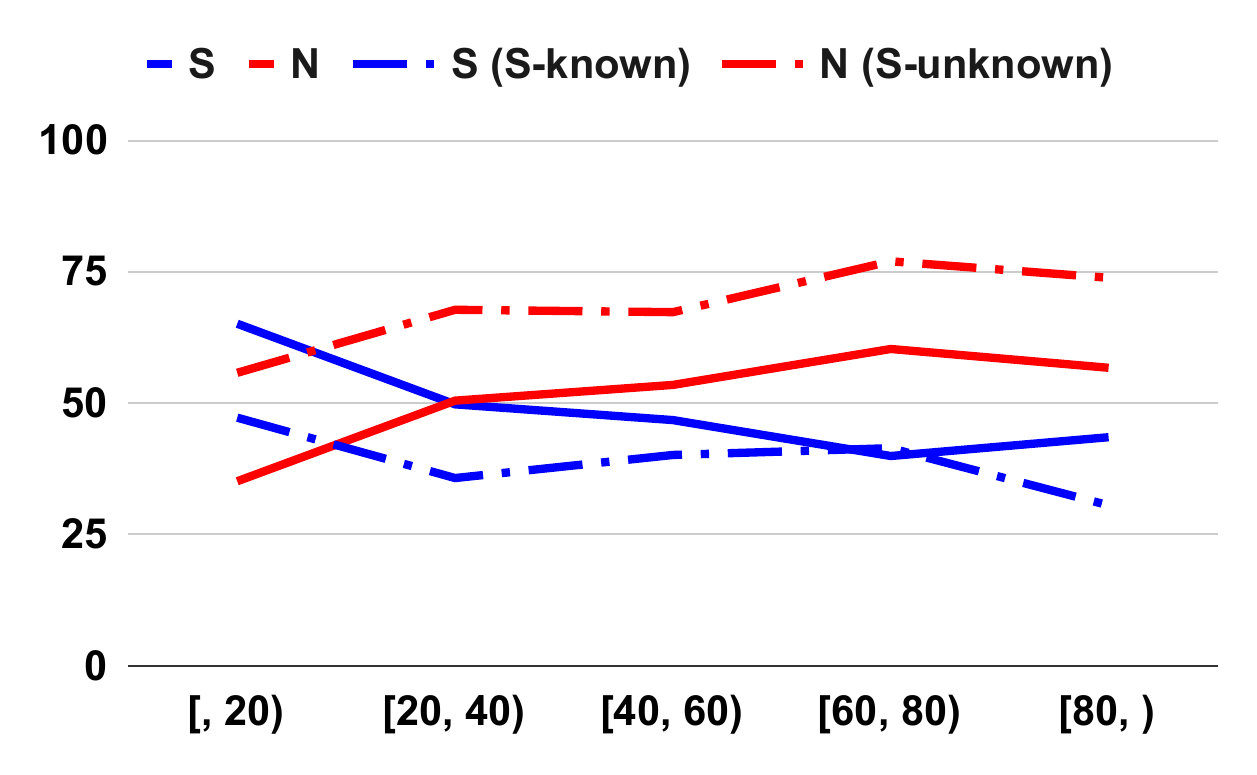}
    \caption{Claude-3-opus}
  \end{subfigure}
  \begin{subfigure}[b]{0.45\linewidth}
    \includegraphics[width=\linewidth]{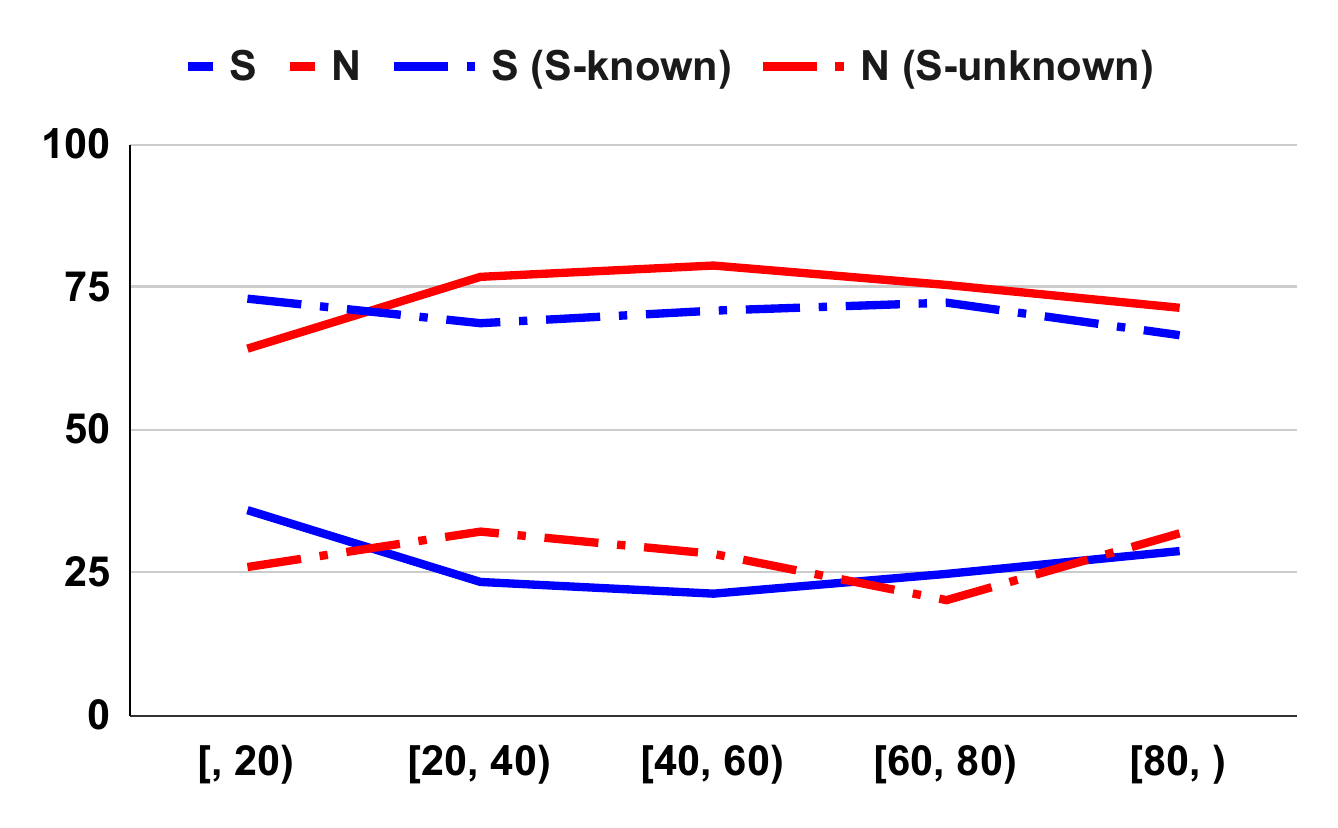}
    \caption{Mixtral-8x7b}
  \end{subfigure}
    \begin{subfigure}[b]{0.45\linewidth}
    \includegraphics[width=\linewidth]{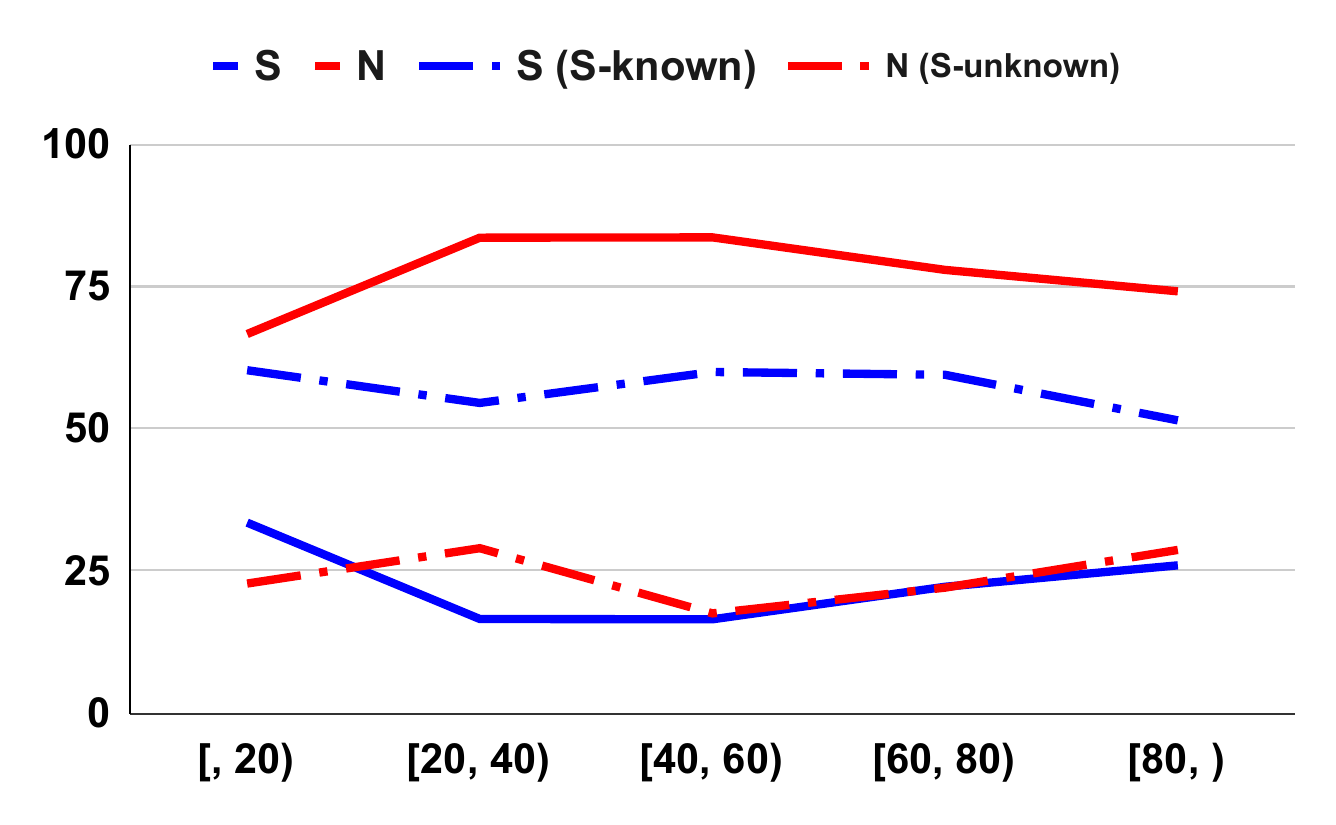}
    \caption{Mistral-Large}
  \end{subfigure}
  
\caption[Two numerical solutions]{Self-Know and Self-Unknown results of different LLMs across different relative positions (\%). \textbf{S}: \textbf{factuality} (percentage of supported atomic claims); \textbf{N}: percentage of unsupported atomic claims; \textbf{S (S-known)}: \textbf{Self-Known} score; \textbf{N (S-unknown)}: \textbf{Self-Unknown} score.  
}
\label{fig:relative_position_llms}
\end{center}
\end{figure*}

\subsubsection{Results}


Figure~\ref{fig:relative_position_llms} show results of several powerful LLMs ( Gemini-1.5-pro, Claude-3-opus, and two Mistral AI models). Two additional LLMs (GPT-4, and Llama-3-70B-Instruct) results are provided in Figure~\ref{fig:relative_position_llms_more} in the Appendix. 

\paragraph{Decreasing Factuality: Strong Start, Later Decline} According to the bold blue lines in Figure~\ref{fig:relative_position_llms}, we observe the highest factuality scores are observed at the beginning of the generated text across all relative positions. 

\paragraph{Factuality Vs. Self-Known Vs. Self-Unknown}  Overall, we observe that the Self-Known score is positively correlated with factuality, as indicated by the \textbf{two blue lines}, and the Self-Unknown score is positively correlated with the percentage of unsupported atomic claims, as shown by the \textbf{two red lines} in each figure. For these advanced LLMs, the trend of these three scores across different positions shows smaller variation.  

\begin{figure}[h]
 \begin{center}
  \includegraphics[width=0.9\columnwidth]{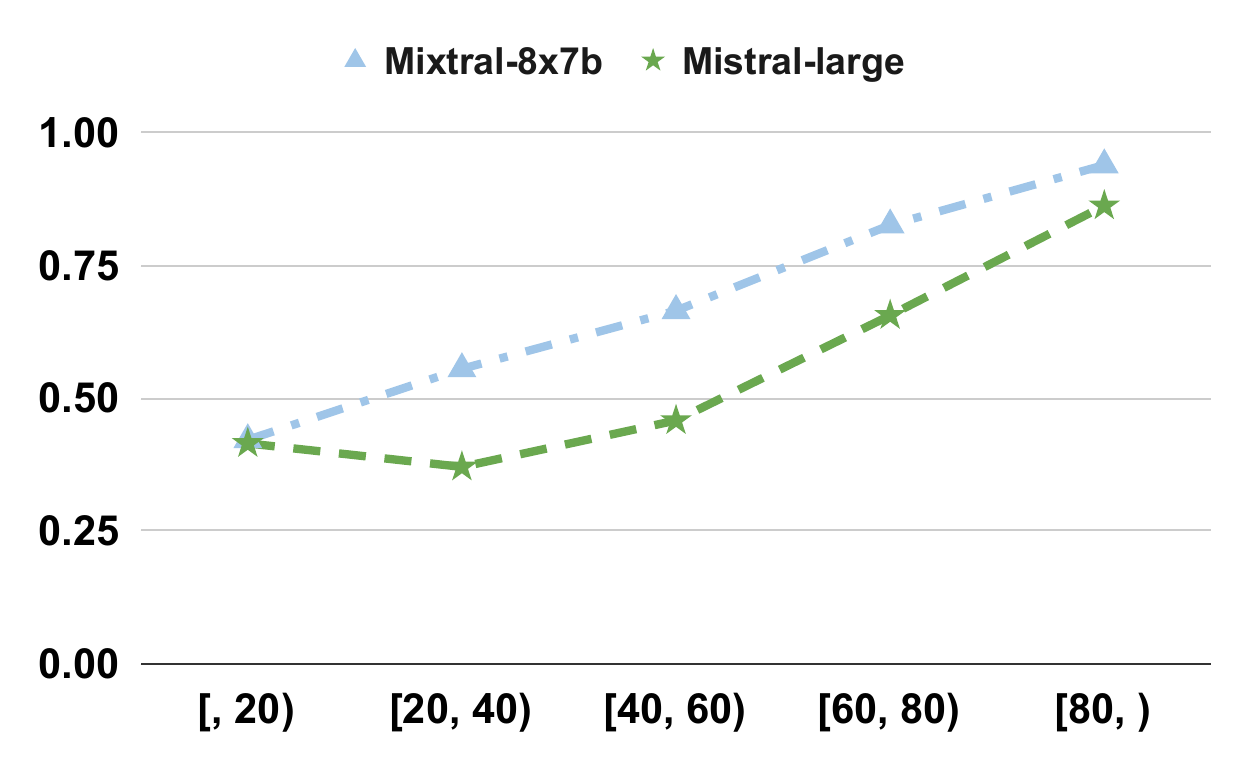}
  \caption{There may be minimal change in the factuality score, but a significant increase in the number of unsupported claims across different relative positions(\%).}
  \label{fig:factEstimationNum}
\end{center}
\end{figure}

\paragraph{Clear Difference in the Number of Unsupported Claims Across Positions} In Figure~\ref{fig:relative_position_llms} (e) and (f), observed minimal differences in factuality for the two models (Mixtral-8x7b and Mistral-Large). However, as depicted in Figure~\ref{fig:factEstimationNum}, the number of unsupported claims increases significantly from the beginning to the end of the generated text. It indicates the challenges of long-form generation. This also highlights a limitation in relying solely on factuality scores for evaluation.

\paragraph{No Significant Changes in Self-Judgment for Some Advanced LLMs} We can observe that there is no big change according to dashed lines (Self-Known and Self-Unknown) in Figure~\ref{fig:relative_position_llms}. However, the number of unsupported claims are increasing as shown in Figure~\ref{fig:factEstimationNum}.


\paragraph{How to Improve Factuality Score?} In Equation~\ref{eq:factscore}, we propose estimating the factuality of a LLM using Self-Known and Self-Unknown scores. A higher Self-Known score typically corresponds to higher factuality. However, does this mean LLMs would achieve 100\% factuality if they had a 100\% Self-Known score and 0 Self-Unknown score on their own generation? The answer is no. It is a necessary condition, not a sufficient one for achieving 100\% factuality. In the derivation of Equation~\ref{eq:factscore}, several additional assumptions are made\footnote{For instance, one key assumption is that the probability of correctness given the model output and input $P (d=1 \mid \y, \x)$, equals the factuality score $\sigma$ of output $\y$, However, if a LLM becomes overconfident in generating answers, the term $P (d=1 \mid \y, \x)$ may significantly exceed the actual factuality score.}. 

According to our results, a higher Self-Known score is usually associated with higher factuality, while a higher Self-Unknown score is associated with lower factuality for LLMs. 
This indicates that it is challenging for LLMs to recognize unsupported claims on their own. Therefore, a judgment model that incorporates an external knowledge source is necessary for this recognition.



Some reasonable questions arise: Are decoding errors in LLMs caused by a lack of relevant knowledge? Can Retrieval-Augmented Generation (RAG), which supplies additional context, address the decline in factuality during later stages of generation? In the next section, we present our exploration of RAG-based experiments across different LLMs.



\subsection{Retrieval-Augmented Generation}
Retrieval-Augmented Generation (RAG) is a widely used approach for enhancing language model performance in various applications. In RAG, relevant text segments are retrieved from an external knowledge source and integrated into the model's responses. For our retrieval corpus, we utilized the English Wikipedia as of April 1, 2023, with each page divided into chunks of up to 256 tokens. These retrieved passages, containing facts relevant to the entity, were incorporated into the LLMs' context to improve the factual accuracy of the generated content.\footnote{One example is shown in Table~\ref{tab:ragExamples}.}.

\begin{figure*}[h!]
\begin{center}
    
\begin{subfigure}[b]{0.45\linewidth}
    \includegraphics[width=\linewidth]{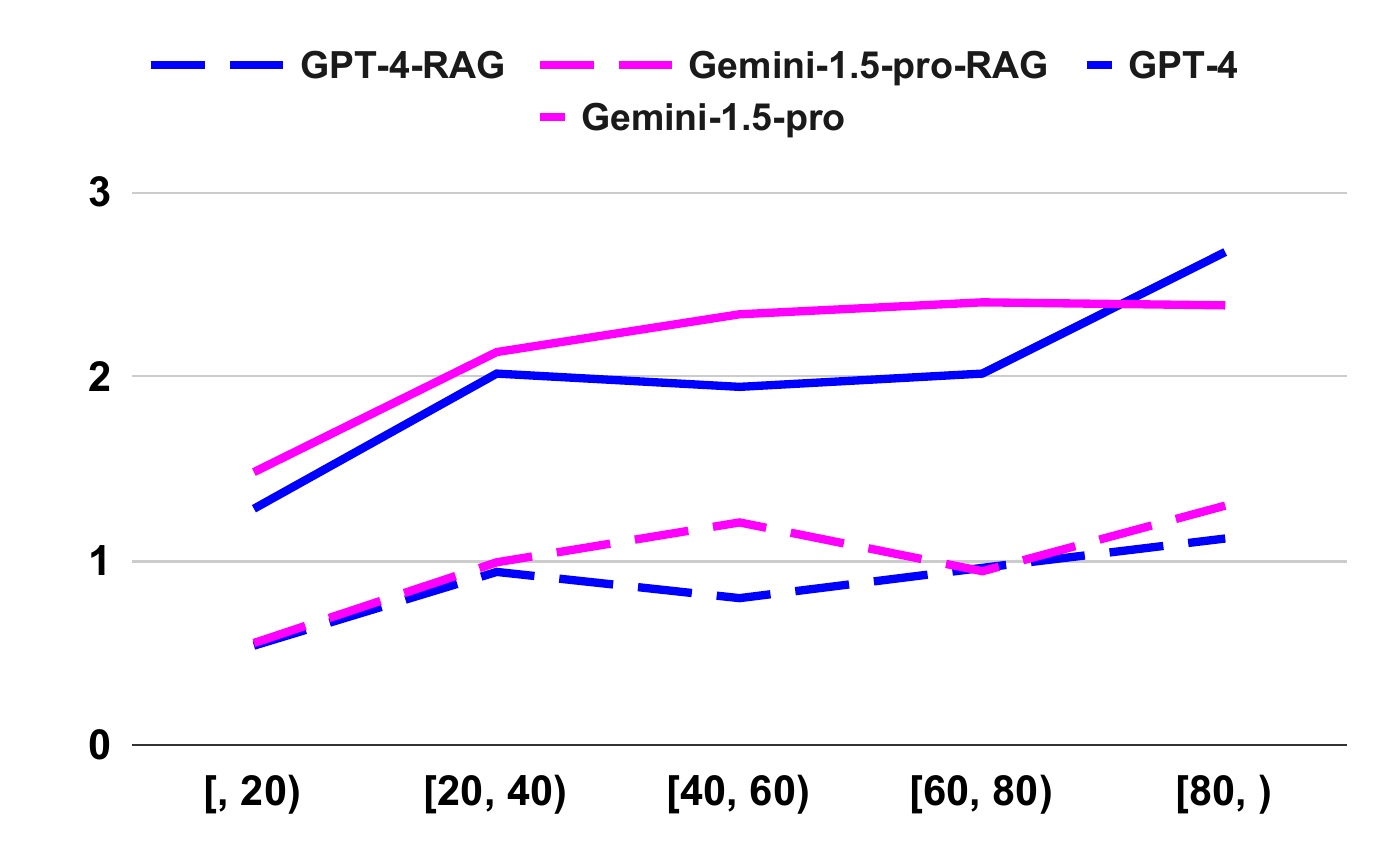}
    \caption{Number of unsupported claims for two LLMs and their RAG models across different relative positions(\%).}
    \label{fig:RAGfactEstimationNum}
  \end{subfigure}\hfill
    \begin{subfigure}[b]{0.45\linewidth}
    \includegraphics[width=\linewidth]{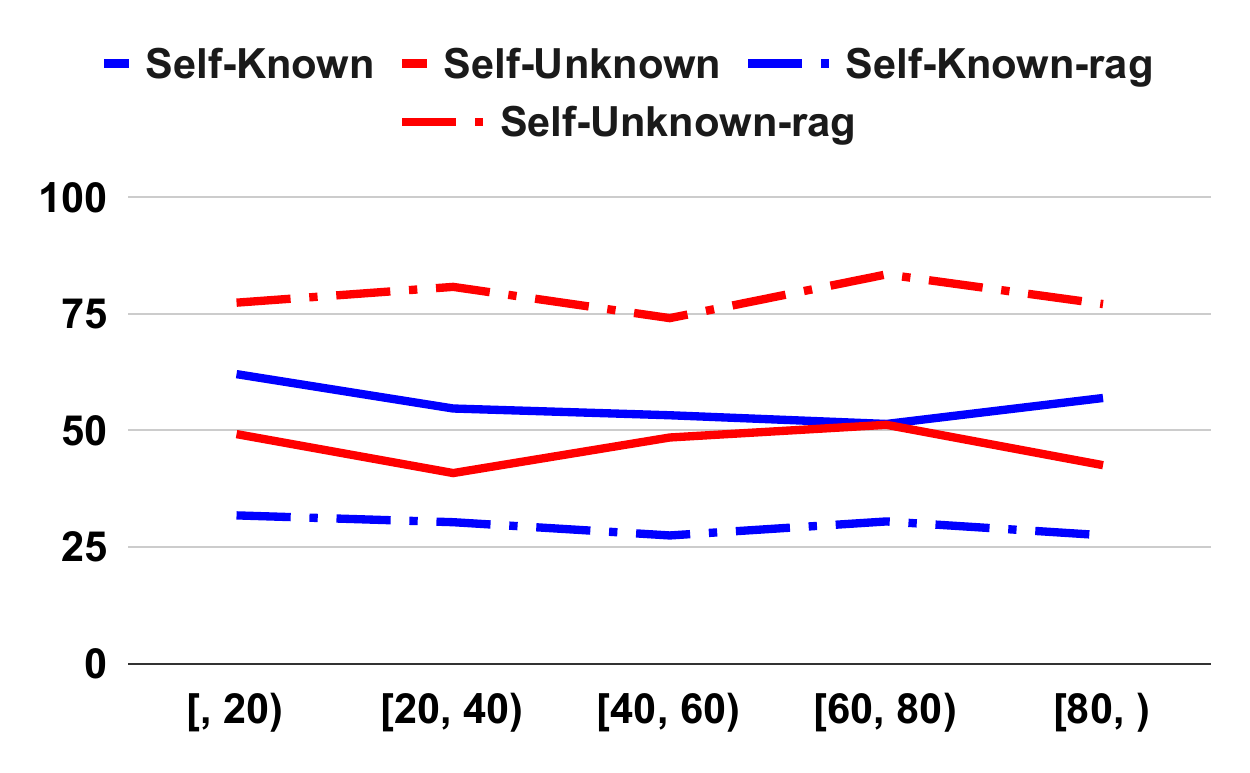}
    \caption{Self-Known and Self-Unknown scores for Gemini-1.5-pro and the RAG model across different relative positions(\%).}
    \label{fig:RAGScore}
  \end{subfigure}

\end{center}
\caption{RAG experiments on two strong LLMs (Gemini-1.5-pro and GPT-4).}
\label{fig:RAG}
\end{figure*}


According to Figure~\ref{fig:RAG} (a), in the RAG setting, although there are significantly fewer unsupported atomic claims overall, a notable increase in the number of unsupported claims is observed in later stages of generation. As shown in Table~\ref{tab:ragExamples}, LLMs can still response with lots of unsupported claim even given context knowledge. This increase is likely due to error propagation within the LLMs, highlighting the challenges of long-form generation even when relevant parts are provided.


Figure~\ref{fig:RAG} (b) demonstrates that the RAG system exhibits significantly lower Self-Known scores and higher Self-Unknown scores. This discrepancy may stem from the corresponding LLM's lack of prior knowledge regarding the retrieved content in the RAG system, causing it to mistakenly assess accurate information as incorrect.

In these RAG experiments, incorporating all relevant knowledge leads to improved factuality in LLMs. However, they still exhibit a decline in factuality during later stages of generation. This underscores the need for alternative frameworks specifically tailored to long-form generation tasks. For instance, employing more sophisticated decoding strategies may help mitigate the challenges associated with long-form generation.



\section{Related Work}
\paragraph{Factuality Evaluation}

Recent advancements have seen significant efforts in quantifying the factuality of LLM generations. For short answers, factuality often correlates with fact verification, which directly assesses whether the generation aligns with extensive knowledge sources and references \citep{thorne-etal-2018-fever, honovich-etal-2022-true-evaluating} or utilizes language models \citep{lin-etal-2022-truthfulqa}. However, evaluating factuality in long-form content poses greater challenges due to the complexity of the generation process. Recent studies \citep{fan-etal-2020-generating, wright-etal-2022-generating, min-etal-2023-factscore} have approached this challenge by breaking down long generations into atomic claims. While these approaches predominantly focus on factual precision, some studies \citep{wei2024long} also consider evaluating factual recall. In our work, we concentrate on factual precision akin to \citet{min-etal-2023-factscore}. Moving forward, the development of more robust automatic tools will be crucial for advancing factuality exploration in long-form generation tasks.

\paragraph{Self-Know and Self-Unknown}


Recent studies have extensively explored the concepts of Self-Known and Self-Unknown in language models. For instance, \citet{kadavath2022language,liu-etal-2022-token,guerreiro-etal-2023-looking} demonstrated that language models are capable of assessing the validity of their own claims and predicting their ability with answering true/false questions accurately. Meanwhile, \citet{srivastava2023imitation,yin-etal-2023-large} highlighted the limitations of LLMs in acknowledging their unknowns, focusing on their ability to recognize unknown knowledge. In our work, we specifically investigate whether LLMs can identify and reconsider unsupported claims generated from their own outputs. Our results indicate that LLMs struggle to accurately judge unsupported atomic claims from their own generations. We also find that a lower Self-Unknown score or a higher Self-Known score corresponds to higher factuality.


\section{Conclusion}
n this study, we investigate the factuality of long-form text generation across different language model families and at various stages of generation. We observe a consistent decline in factuality in sentences generated later in the sequence. To understand the underlying causes, we explore two possible factors: diminished self-knowledge in later generations and the accumulation of earlier generation errors (i.e., error propagation). To analyze this, we introduce the concepts of Self-Known and Self-Unknown scores, which measure a model's confidence in its own knowledge. We find that current LLMs struggle to maintain factual accuracy over extended generations, partly due to limitations in their internal knowledge representation and propagation mechanisms. Addressing these challenges requires further research. Promising directions include the development of external factuality evaluation modules (e.g., dedicated judge models) and the design of more robust decoding strategies tailored to long-form generation





\section{Limitations}
Following are limitations in our work.

\paragraph{Evaluation of Self-Know and Self-Unknown}
In this work, we design three different methods for estimating Self-Known and Self-Unknown scores on LLMs' own generation. We find that the third setting (c), which includes the option "None of the above," is effective in determining whether LLMs can accurately judge the correctness of claims generated from their own outputs. Although our results show that these scores are well aligned with the estimation of factuality scores using Equation~\ref{eq:factscore}, exploring better methods for evaluating the correctness of claims with LLMs would still be beneficial for future study.

\paragraph{Factuality Evaluation} In this work, we limit the domain of long-form generation to ensure accurate factuality evaluation. The concern is that broadening the topic range might compromise the accuracy of our factuality assessments, rendering our study less effective. Recently, evaluation tools~\citep{guan2024language,es2023ragas,tang2024minicheck,wei2024long} have been explored. With stronger tools, it is possible to explore a wider range of domains beyond Wikipedia.

Moreover, in this work, we primarily focus on factuality precision. However, considering factuality recall is also important, as it ensures that the omission of significant pieces of information is penalized during evaluation. By incorporating both precision and recall, we can achieve a more comprehensive and accurate assessment of factuality in long-form generation.

\bibliography{conference}

\begin{thebibliography}{35}
\providecommand{\natexlab}[1]{#1}

\bibitem[{AI@Meta(2024)}]{llama3modelcard}
AI@Meta. 2024.
\newblock \href {https://github.com/meta-llama/llama3/blob/main/MODEL_CARD.md} {Llama 3 model card}.

\bibitem[{An et~al.(2024)An, Gong, Zhong, Zhao, Li, Zhang, Kong, and Qiu}]{an-etal-2024-l}
Chenxin An, Shansan Gong, Ming Zhong, Xingjian Zhao, Mukai Li, Jun Zhang, Lingpeng Kong, and Xipeng Qiu. 2024.
\newblock \href {https://doi.org/10.18653/v1/2024.acl-long.776} {{L}-eval: Instituting standardized evaluation for long context language models}.
\newblock In \emph{Proceedings of the 62nd Annual Meeting of the Association for Computational Linguistics (Volume 1: Long Papers)}, pages 14388--14411, Bangkok, Thailand. Association for Computational Linguistics.

\bibitem[{Anthropic(2024)}]{claude3}
Anthropic. 2024.
\newblock \href {https://www.anthropic.com/news/claude-3-family.} {Introducing the next generation of claude}.

\bibitem[{Bai et~al.(2024)Bai, Lv, Zhang, Lyu, Tang, Huang, Du, Liu, Zeng, Hou, Dong, Tang, and Li}]{bai-etal-2024-longbench}
Yushi Bai, Xin Lv, Jiajie Zhang, Hongchang Lyu, Jiankai Tang, Zhidian Huang, Zhengxiao Du, Xiao Liu, Aohan Zeng, Lei Hou, Yuxiao Dong, Jie Tang, and Juanzi Li. 2024.
\newblock \href {https://doi.org/10.18653/v1/2024.acl-long.172} {{L}ong{B}ench: A bilingual, multitask benchmark for long context understanding}.
\newblock In \emph{Proceedings of the 62nd Annual Meeting of the Association for Computational Linguistics (Volume 1: Long Papers)}, pages 3119--3137, Bangkok, Thailand. Association for Computational Linguistics.

\bibitem[{Chen et~al.(2022)Chen, Sriram, Choi, and Durrett}]{chen-etal-2022-generating}
Jifan Chen, Aniruddh Sriram, Eunsol Choi, and Greg Durrett. 2022.
\newblock \href {https://doi.org/10.18653/v1/2022.emnlp-main.229} {Generating literal and implied subquestions to fact-check complex claims}.
\newblock In \emph{Proceedings of the 2022 Conference on Empirical Methods in Natural Language Processing}, pages 3495--3516, Abu Dhabi, United Arab Emirates. Association for Computational Linguistics.

\bibitem[{Es et~al.(2023)Es, James, Espinosa-Anke, and Schockaert}]{es2023ragas}
Shahul Es, Jithin James, Luis Espinosa-Anke, and Steven Schockaert. 2023.
\newblock \href {https://arxiv.org/abs/2309.15217} {Ragas: Automated evaluation of retrieval augmented generation}.
\newblock \emph{Preprint}, arXiv:2309.15217.

\bibitem[{Fan et~al.(2020)Fan, Piktus, Petroni, Wenzek, Saeidi, Vlachos, Bordes, and Riedel}]{fan-etal-2020-generating}
Angela Fan, Aleksandra Piktus, Fabio Petroni, Guillaume Wenzek, Marzieh Saeidi, Andreas Vlachos, Antoine Bordes, and Sebastian Riedel. 2020.
\newblock \href {https://doi.org/10.18653/v1/2020.emnlp-main.580} {Generating fact checking briefs}.
\newblock In \emph{Proceedings of the 2020 Conference on Empirical Methods in Natural Language Processing (EMNLP)}, pages 7147--7161, Online. Association for Computational Linguistics.

\bibitem[{GeminiTeam(2024)}]{geminiteam2024gemini}
GeminiTeam. 2024.
\newblock \href {https://arxiv.org/abs/2403.05530} {Gemini 1.5: Unlocking multimodal understanding across millions of tokens of context}.
\newblock \emph{Preprint}, arXiv:2403.05530.

\bibitem[{Guan et~al.(2024)Guan, Dodge, Wadden, Huang, and Peng}]{guan2024language}
Jian Guan, Jesse Dodge, David Wadden, Minlie Huang, and Hao Peng. 2024.
\newblock \href {https://arxiv.org/abs/2310.14564} {Language models hallucinate, but may excel at fact verification}.
\newblock \emph{Preprint}, arXiv:2310.14564.

\bibitem[{Guerreiro et~al.(2023)Guerreiro, Voita, and Martins}]{guerreiro-etal-2023-looking}
Nuno~M. Guerreiro, Elena Voita, and Andr{\'e} Martins. 2023.
\newblock \href {https://doi.org/10.18653/v1/2023.eacl-main.75} {Looking for a needle in a haystack: A comprehensive study of hallucinations in neural machine translation}.
\newblock In \emph{Proceedings of the 17th Conference of the European Chapter of the Association for Computational Linguistics}, pages 1059--1075, Dubrovnik, Croatia. Association for Computational Linguistics.

\bibitem[{Honovich et~al.(2022)Honovich, Aharoni, Herzig, Taitelbaum, Kukliansy, Cohen, Scialom, Szpektor, Hassidim, and Matias}]{honovich-etal-2022-true-evaluating}
Or~Honovich, Roee Aharoni, Jonathan Herzig, Hagai Taitelbaum, Doron Kukliansy, Vered Cohen, Thomas Scialom, Idan Szpektor, Avinatan Hassidim, and Yossi Matias. 2022.
\newblock \href {https://doi.org/10.18653/v1/2022.naacl-main.287} {{TRUE}: Re-evaluating factual consistency evaluation}.
\newblock In \emph{Proceedings of the 2022 Conference of the North American Chapter of the Association for Computational Linguistics: Human Language Technologies}, pages 3905--3920, Seattle, United States. Association for Computational Linguistics.

\bibitem[{Hu et~al.(2024)Hu, Chen, Li, Guo, Wen, Yu, and Guo}]{hu2024towards}
Xuming Hu, Junzhe Chen, Xiaochuan Li, Yufei Guo, Lijie Wen, Philip~S. Yu, and Zhijiang Guo. 2024.
\newblock \href {https://openreview.net/forum?id=9OevMUdods} {Towards understanding factual knowledge of large language models}.
\newblock In \emph{The Twelfth International Conference on Learning Representations}.

\bibitem[{Jiang et~al.(2024)Jiang, Sablayrolles, Roux, Mensch, Savary, Bamford, Chaplot, de~las Casas, Hanna, Bressand, Lengyel, Bour, Lample, Lavaud, Saulnier, Lachaux, Stock, Subramanian, Yang, Antoniak, Scao, Gervet, Lavril, Wang, Lacroix, and Sayed}]{jiang2024mixtral}
Albert~Q. Jiang, Alexandre Sablayrolles, Antoine Roux, Arthur Mensch, Blanche Savary, Chris Bamford, Devendra~Singh Chaplot, Diego de~las Casas, Emma~Bou Hanna, Florian Bressand, Gianna Lengyel, Guillaume Bour, Guillaume Lample, Lélio~Renard Lavaud, Lucile Saulnier, Marie-Anne Lachaux, Pierre Stock, Sandeep Subramanian, Sophia Yang, Szymon Antoniak, Teven~Le Scao, Théophile Gervet, Thibaut Lavril, Thomas Wang, Timothée Lacroix, and William~El Sayed. 2024.
\newblock \href {https://arxiv.org/abs/2401.04088} {Mixtral of experts}.
\newblock \emph{Preprint}, arXiv:2401.04088.

\bibitem[{Kadavath et~al.(2022)Kadavath, Conerly, Askell, Henighan, Drain, Perez, Schiefer, Hatfield-Dodds, DasSarma, Tran-Johnson, Johnston, El-Showk, Jones, Elhage, Hume, Chen, Bai, Bowman, Fort, Ganguli, Hernandez, Jacobson, Kernion, Kravec, Lovitt, Ndousse, Olsson, Ringer, Amodei, Brown, Clark, Joseph, Mann, McCandlish, Olah, and Kaplan}]{kadavath2022language}
Saurav Kadavath, Tom Conerly, Amanda Askell, Tom Henighan, Dawn Drain, Ethan Perez, Nicholas Schiefer, Zac Hatfield-Dodds, Nova DasSarma, Eli Tran-Johnson, Scott Johnston, Sheer El-Showk, Andy Jones, Nelson Elhage, Tristan Hume, Anna Chen, Yuntao Bai, Sam Bowman, Stanislav Fort, Deep Ganguli, Danny Hernandez, Josh Jacobson, Jackson Kernion, Shauna Kravec, Liane Lovitt, Kamal Ndousse, Catherine Olsson, Sam Ringer, Dario Amodei, Tom Brown, Jack Clark, Nicholas Joseph, Ben Mann, Sam McCandlish, Chris Olah, and Jared Kaplan. 2022.
\newblock \href {https://arxiv.org/abs/2207.05221} {Language models (mostly) know what they know}.
\newblock \emph{Preprint}, arXiv:2207.05221.

\bibitem[{Kuratov et~al.(2024)Kuratov, Bulatov, Anokhin, Rodkin, Sorokin, Sorokin, and Burtsev}]{kuratov2024babilongtestinglimitsllms}
Yuri Kuratov, Aydar Bulatov, Petr Anokhin, Ivan Rodkin, Dmitry Sorokin, Artyom Sorokin, and Mikhail Burtsev. 2024.
\newblock \href {https://arxiv.org/abs/2406.10149} {Babilong: Testing the limits of llms with long context reasoning-in-a-haystack}.
\newblock \emph{Preprint}, arXiv:2406.10149.

\bibitem[{Li et~al.(2023)Li, Cheng, Zhao, Nie, and Wen}]{li-etal-2023-halueval}
Junyi Li, Xiaoxue Cheng, Xin Zhao, Jian-Yun Nie, and Ji-Rong Wen. 2023.
\newblock \href {https://doi.org/10.18653/v1/2023.emnlp-main.397} {{H}alu{E}val: A large-scale hallucination evaluation benchmark for large language models}.
\newblock In \emph{Proceedings of the 2023 Conference on Empirical Methods in Natural Language Processing}, pages 6449--6464, Singapore. Association for Computational Linguistics.

\bibitem[{Lin et~al.(2022)Lin, Hilton, and Evans}]{lin-etal-2022-truthfulqa}
Stephanie Lin, Jacob Hilton, and Owain Evans. 2022.
\newblock \href {https://doi.org/10.18653/v1/2022.acl-long.229} {{T}ruthful{QA}: Measuring how models mimic human falsehoods}.
\newblock In \emph{Proceedings of the 60th Annual Meeting of the Association for Computational Linguistics (Volume 1: Long Papers)}, pages 3214--3252, Dublin, Ireland. Association for Computational Linguistics.

\bibitem[{Liu et~al.(2022)Liu, Zhang, Brockett, Mao, Sui, Chen, and Dolan}]{liu-etal-2022-token}
Tianyu Liu, Yizhe Zhang, Chris Brockett, Yi~Mao, Zhifang Sui, Weizhu Chen, and Bill Dolan. 2022.
\newblock \href {https://doi.org/10.18653/v1/2022.acl-long.464} {A token-level reference-free hallucination detection benchmark for free-form text generation}.
\newblock In \emph{Proceedings of the 60th Annual Meeting of the Association for Computational Linguistics (Volume 1: Long Papers)}, pages 6723--6737, Dublin, Ireland. Association for Computational Linguistics.

\bibitem[{Liu et~al.(2023)Liu, Fabbri, Liu, Zhao, Nan, Han, Han, Joty, Wu, Xiong, and Radev}]{liu-etal-2023-revisiting}
Yixin Liu, Alex Fabbri, Pengfei Liu, Yilun Zhao, Linyong Nan, Ruilin Han, Simeng Han, Shafiq Joty, Chien-Sheng Wu, Caiming Xiong, and Dragomir Radev. 2023.
\newblock \href {https://doi.org/10.18653/v1/2023.acl-long.228} {Revisiting the gold standard: Grounding summarization evaluation with robust human evaluation}.
\newblock In \emph{Proceedings of the 61st Annual Meeting of the Association for Computational Linguistics (Volume 1: Long Papers)}, pages 4140--4170, Toronto, Canada. Association for Computational Linguistics.

\bibitem[{Manakul et~al.(2023)Manakul, Liusie, and Gales}]{manakul-etal-2023-selfcheckgpt}
Potsawee Manakul, Adian Liusie, and Mark Gales. 2023.
\newblock \href {https://doi.org/10.18653/v1/2023.emnlp-main.557} {{S}elf{C}heck{GPT}: Zero-resource black-box hallucination detection for generative large language models}.
\newblock In \emph{Proceedings of the 2023 Conference on Empirical Methods in Natural Language Processing}, pages 9004--9017, Singapore. Association for Computational Linguistics.

\bibitem[{Min et~al.(2023)Min, Krishna, Lyu, Lewis, Yih, Koh, Iyyer, Zettlemoyer, and Hajishirzi}]{min-etal-2023-factscore}
Sewon Min, Kalpesh Krishna, Xinxi Lyu, Mike Lewis, Wen-tau Yih, Pang Koh, Mohit Iyyer, Luke Zettlemoyer, and Hannaneh Hajishirzi. 2023.
\newblock \href {https://doi.org/10.18653/v1/2023.emnlp-main.741} {{FA}ct{S}core: Fine-grained atomic evaluation of factual precision in long form text generation}.
\newblock In \emph{Proceedings of the 2023 Conference on Empirical Methods in Natural Language Processing}, pages 12076--12100, Singapore. Association for Computational Linguistics.

\bibitem[{OpenAI(2023{\natexlab{a}})}]{openai2023gpt3.5}
OpenAI. 2023{\natexlab{a}}.
\newblock Chatgpt.
\newblock \emph{arXiv preprint arXiv:2303.08774}.

\bibitem[{OpenAI(2023{\natexlab{b}})}]{openai2023gpt4}
OpenAI. 2023{\natexlab{b}}.
\newblock Gpt-4 technical report.
\newblock \emph{arXiv preprint arXiv:2303.08774}.

\bibitem[{Ouyang et~al.(2022)Ouyang, Wu, Jiang, Almeida, Wainwright, Mishkin, Zhang, Agarwal, Slama, Ray, Schulman, Hilton, Kelton, Miller, Simens, Askell, Welinder, Christiano, Leike, and Lowe}]{NEURIPS2022_b1efde53}
Long Ouyang, Jeffrey Wu, Xu~Jiang, Diogo Almeida, Carroll Wainwright, Pamela Mishkin, Chong Zhang, Sandhini Agarwal, Katarina Slama, Alex Ray, John Schulman, Jacob Hilton, Fraser Kelton, Luke Miller, Maddie Simens, Amanda Askell, Peter Welinder, Paul~F Christiano, Jan Leike, and Ryan Lowe. 2022.
\newblock \href {https://proceedings.neurips.cc/paper_files/paper/2022/file/b1efde53be364a73914f58805a001731-Paper-Conference.pdf} {Training language models to follow instructions with human feedback}.
\newblock In \emph{Advances in Neural Information Processing Systems}, volume~35, pages 27730--27744. Curran Associates, Inc.

\bibitem[{Rajpurkar et~al.(2018)Rajpurkar, Jia, and Liang}]{rajpurkar-etal-2018-know}
Pranav Rajpurkar, Robin Jia, and Percy Liang. 2018.
\newblock \href {https://doi.org/10.18653/v1/P18-2124} {Know what you don{'}t know: Unanswerable questions for {SQ}u{AD}}.
\newblock In \emph{Proceedings of the 56th Annual Meeting of the Association for Computational Linguistics (Volume 2: Short Papers)}, pages 784--789, Melbourne, Australia. Association for Computational Linguistics.

\bibitem[{Shaham et~al.(2023)Shaham, Ivgi, Efrat, Berant, and Levy}]{shaham-etal-2023-zeroscrolls}
Uri Shaham, Maor Ivgi, Avia Efrat, Jonathan Berant, and Omer Levy. 2023.
\newblock \href {https://doi.org/10.18653/v1/2023.findings-emnlp.536} {{Z}ero{SCROLLS}: A zero-shot benchmark for long text understanding}.
\newblock In \emph{Findings of the Association for Computational Linguistics: EMNLP 2023}, pages 7977--7989, Singapore. Association for Computational Linguistics.

\bibitem[{Srivastava~et(2023)}]{srivastava2023imitation}
al~Srivastava~et. 2023.
\newblock \href {https://arxiv.org/abs/2206.04615} {Beyond the imitation game: Quantifying and extrapolating the capabilities of language models}.
\newblock \emph{Preprint}, arXiv:2206.04615.

\bibitem[{Tang et~al.(2024)Tang, Laban, and Durrett}]{tang2024minicheck}
Liyan Tang, Philippe Laban, and Greg Durrett. 2024.
\newblock \href {https://arxiv.org/abs/2404.10774} {Minicheck: Efficient fact-checking of llms on grounding documents}.
\newblock \emph{Preprint}, arXiv:2404.10774.

\bibitem[{Thorne et~al.(2018)Thorne, Vlachos, Christodoulopoulos, and Mittal}]{thorne-etal-2018-fever}
James Thorne, Andreas Vlachos, Christos Christodoulopoulos, and Arpit Mittal. 2018.
\newblock \href {https://doi.org/10.18653/v1/N18-1074} {{FEVER}: a large-scale dataset for fact extraction and {VER}ification}.
\newblock In \emph{Proceedings of the 2018 Conference of the North {A}merican Chapter of the Association for Computational Linguistics: Human Language Technologies, Volume 1 (Long Papers)}, pages 809--819, New Orleans, Louisiana. Association for Computational Linguistics.

\bibitem[{Trischler et~al.(2017)Trischler, Wang, Yuan, Harris, Sordoni, Bachman, and Suleman}]{trischler-etal-2017-newsqa}
Adam Trischler, Tong Wang, Xingdi Yuan, Justin Harris, Alessandro Sordoni, Philip Bachman, and Kaheer Suleman. 2017.
\newblock \href {https://doi.org/10.18653/v1/W17-2623} {{N}ews{QA}: A machine comprehension dataset}.
\newblock In \emph{Proceedings of the 2nd Workshop on Representation Learning for {NLP}}, pages 191--200, Vancouver, Canada. Association for Computational Linguistics.

\bibitem[{Wei et~al.(2024)Wei, Yang, Song, Lu, Hu, Huang, Tran, Peng, Liu, Huang, Du, and Le}]{wei2024long}
Jerry Wei, Chengrun Yang, Xinying Song, Yifeng Lu, Nathan Hu, Jie Huang, Dustin Tran, Daiyi Peng, Ruibo Liu, Da~Huang, Cosmo Du, and Quoc~V. Le. 2024.
\newblock \href {https://arxiv.org/abs/2403.18802} {Long-form factuality in large language models}.

\bibitem[{Wright et~al.(2022)Wright, Wadden, Lo, Kuehl, Cohan, Augenstein, and Wang}]{wright-etal-2022-generating}
Dustin Wright, David Wadden, Kyle Lo, Bailey Kuehl, Arman Cohan, Isabelle Augenstein, and Lucy~Lu Wang. 2022.
\newblock \href {https://doi.org/10.18653/v1/2022.acl-long.175} {Generating scientific claims for zero-shot scientific fact checking}.
\newblock In \emph{Proceedings of the 60th Annual Meeting of the Association for Computational Linguistics (Volume 1: Long Papers)}, pages 2448--2460, Dublin, Ireland. Association for Computational Linguistics.

\bibitem[{Xiong et~al.(2024)Xiong, Hu, Lu, LI, Fu, He, and Hooi}]{xiong2024can}
Miao Xiong, Zhiyuan Hu, Xinyang Lu, YIFEI LI, Jie Fu, Junxian He, and Bryan Hooi. 2024.
\newblock \href {https://openreview.net/forum?id=gjeQKFxFpZ} {Can {LLM}s express their uncertainty? an empirical evaluation of confidence elicitation in {LLM}s}.
\newblock In \emph{The Twelfth International Conference on Learning Representations}.

\bibitem[{Yin et~al.(2023)Yin, Sun, Guo, Wu, Qiu, and Huang}]{yin-etal-2023-large}
Zhangyue Yin, Qiushi Sun, Qipeng Guo, Jiawen Wu, Xipeng Qiu, and Xuanjing Huang. 2023.
\newblock \href {https://doi.org/10.18653/v1/2023.findings-acl.551} {Do large language models know what they don{'}t know?}
\newblock In \emph{Findings of the Association for Computational Linguistics: ACL 2023}, pages 8653--8665, Toronto, Canada. Association for Computational Linguistics.

\bibitem[{Zhang et~al.(2024)Zhang, Chen, Hu, Xu, Chen, Hao, Han, Thai, Wang, Liu, and Sun}]{zhang-etal-2024-bench}
Xinrong Zhang, Yingfa Chen, Shengding Hu, Zihang Xu, Junhao Chen, Moo Hao, Xu~Han, Zhen Thai, Shuo Wang, Zhiyuan Liu, and Maosong Sun. 2024.
\newblock \href {https://doi.org/10.18653/v1/2024.acl-long.814} {$\infty${B}ench: Extending long context evaluation beyond 100{K} tokens}.
\newblock In \emph{Proceedings of the 62nd Annual Meeting of the Association for Computational Linguistics (Volume 1: Long Papers)}, pages 15262--15277, Bangkok, Thailand. Association for Computational Linguistics.

\end{thebibliography}

\appendix

\section{LLMs}
\label{sec:llms}
Following are the models that are used in our experiments.
\begin{itemize}
    \item InstructGPT (text-davinci-003) ~\citep{NEURIPS2022_b1efde53}:
    \item ChatGPT~\citep{openai2023gpt3.5}  
    \item PerplexityAI\footnote{https://www.perplexity.ai/}
    \item GPT-4, GPT-4-Turbo~\citep{openai2023gpt4}
    \item Claude-3-Opus~\citep{claude3} 
    \item Gemini-1.5-Pro~\citep{geminiteam2024gemini}
    \item Mixtral-8x7b~\citep{jiang2024mixtral}
    \item Mistral-Large\footnote{https://mistral.ai/technology/\#models}
    \item Llama-3-70B-Instruct~\citep{llama3modelcard}
\end{itemize}

\section{Prompt Templates}
\label{sec:appendix}

\subsection{Prompts Self-known and Self-unknown}
\paragraph{Direct-Asking}

Given an atomic claim \texttt{\{claim\}} and person \texttt{\{person\}}, we use the following template:
\begin{quote}
    \texttt{Following is a statement from a bio of \{person\}. Please check whether the statement is correct or wrong according to your knowledge. \\\\ \{claim\}\\Is this statement true or false?
    }
\end{quote}

\paragraph{Question-Answering}

Give a question answer pair \texttt{<\{q\}, \{a\}>} that is derived from an atomic claim, the following template is used to determine whether LLMs consider the proposed answer is correct: 

\begin{quote}
    \texttt{Question: \{q\}\\Proposed Answer: \{a\}\\Is the proposed answer: \\ \- (A) True \\ \-  \- (B) False \\ The proposed answer is:
    }
\end{quote}

\paragraph{ Question-Answering w/ None of the above}

Given the question answer pair \texttt{<\{q\}, \{a\}>} derived from an atomic claim, the following template is used: 

\begin{quote}
    \texttt{Question: \{q\}\\Proposed Answer: 
    \{a\}\\Is the proposed answer: \\ \- (A) True \\ \- \- (B) False \\ \- \- (C) None of the above \\ The proposed answer is:
    }
\end{quote}

\subsection{Prompts for Creating the Question-answer Pair}

 Given an atomic claim \texttt{\{claim\}} of a bio and the person \texttt{\{person\}}, a question-answer pair can be derived with gpt-4 with the following template:

\begin{quote}
    \texttt{Following is a fact from a bio of \{person\}. Please ask a question and provide the answer. The answer is as concise as you can, using a single phrase if possible. The answer is also part of the provided fact. The question and answer is separetd with \#. \\\\ \{claim\}
    }
\end{quote}

\section{Rules for Filtering Generations}
\label{sec:rules}
Following are the rules we find that are useful to filter out  unresponsive generation. 
\begin{quote}
    \texttt{I don't have ...} \\
    \texttt{I do not have ...} \\
    \texttt{I need more information ...} \\
    \texttt{Please provide me ...} \\
    \texttt{Please clarify} \\
    \texttt{I apologize ...} \\
    \texttt{there isn't enough information} \\
    \texttt{Unfortunately, there is no ...} \\
    \texttt{If you can provide more information ...} \\
    \texttt{you could provide more ...} \\
    \texttt{It seems you might ...} \\
    
\end{quote}

\section{Automatic Tool Results}
\begin{figure*}[h!]
\centering
\begin{subfigure}[b]{0.48\textwidth}
   \includegraphics[width=1\linewidth]{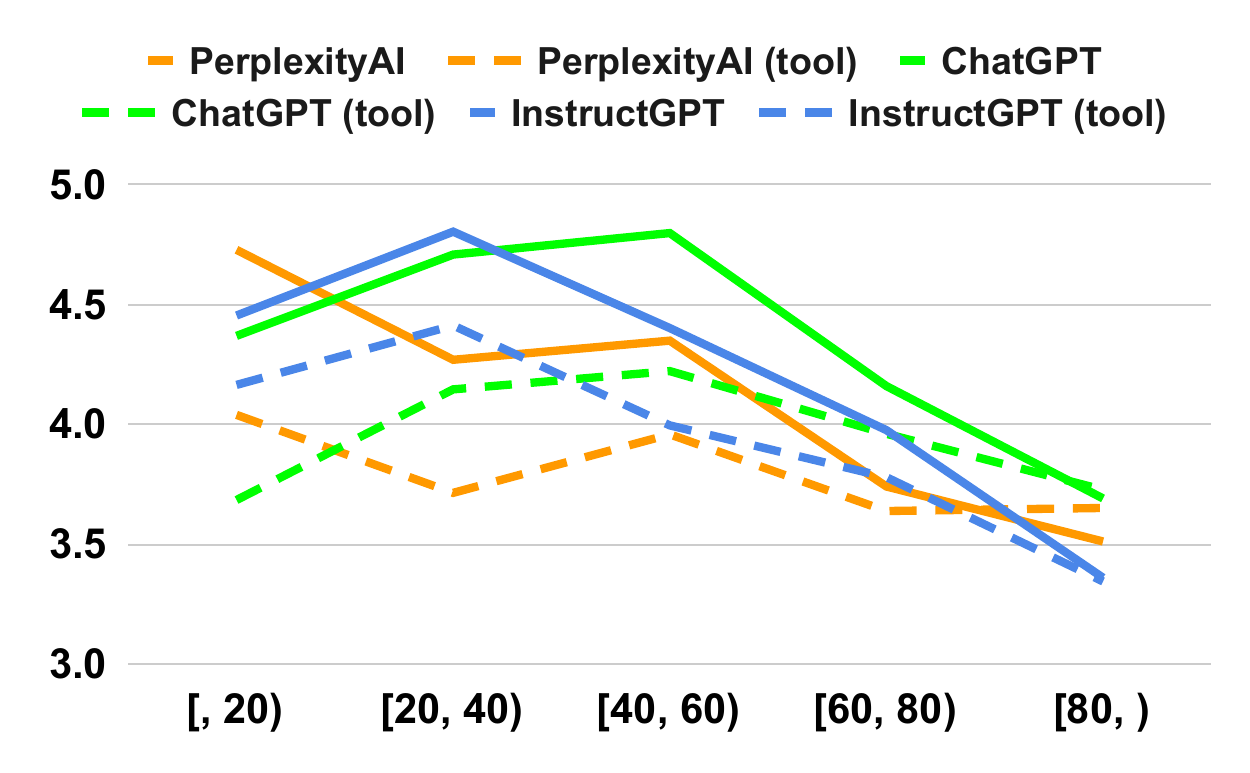}
   \caption{Number of atomic claims}
   \label{fig:Ng1} 
\end{subfigure}

\begin{subfigure}[b]{0.48\textwidth}
   \includegraphics[width=1\linewidth]{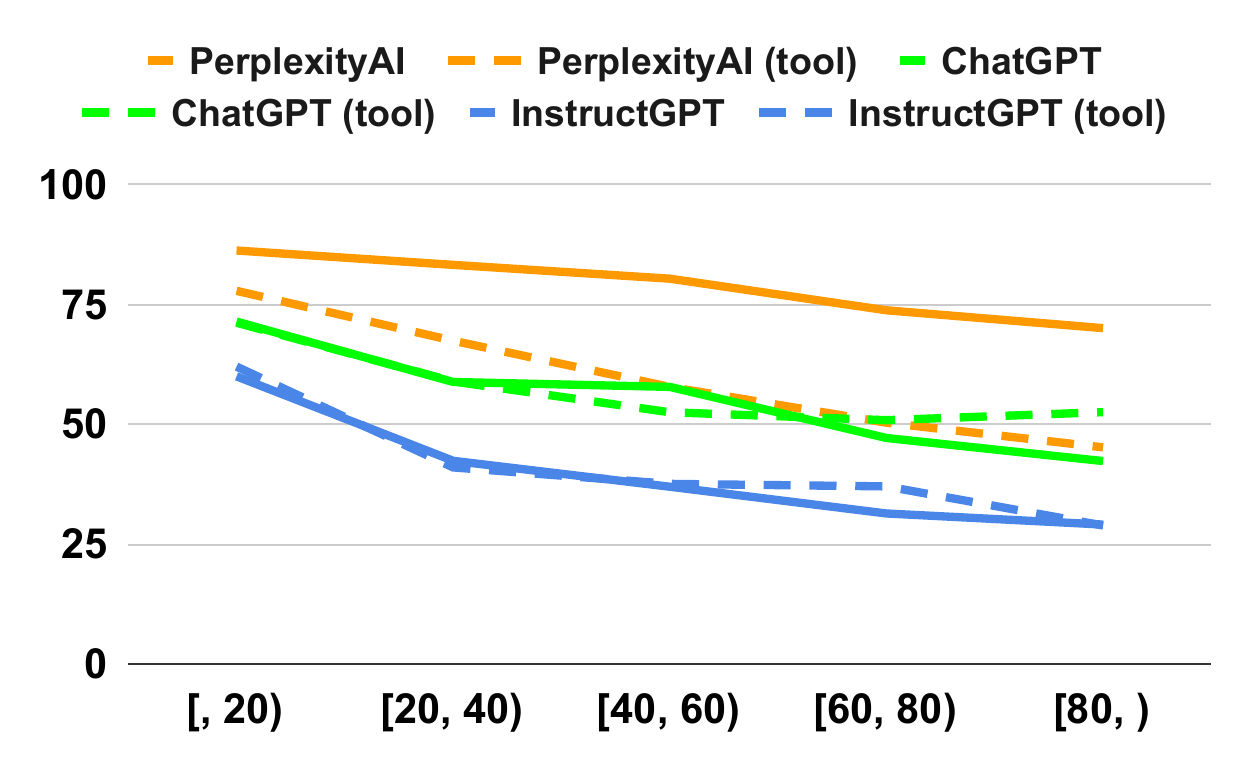}
   \caption{Factuality}
   \label{fig:Ng2}
\end{subfigure}


\caption[Two numerical solutions]{Comparison between our used tool and human annotation. The number of atomic claims and factuality (\%)  across different relative positions (\%) are shown for three LLMs: InstructGPT (text-davinci-003), ChatGPT and PerplexityAI.}
\label{fig:tool_relative_position}
\end{figure*}

See Figure~\ref{fig:tool_relative_position}

\section{Details on Computing Experimental Result For each LLM}
\label{app:steps}
\paragraph{Step 1: Obtaining generations} We feed a prompt ``Tell me a bio of $<$entity$>$'' to the LLM and take the generation. 500 human entities~\citep{min-etal-2023-factscore} are used to generate these biographies. 
\paragraph{Step 2: Filtering generations} For lots of LLMs, a biography is not provided if they think they do not have enough detailed information to provide a biography. We implement rules to filter out these generations\footnote{The useful rules are shown in Section~\ref{sec:rules}. }. 
\paragraph{Step 3: Evaluation factuality} We use the tool for breaking generations into atomic claims and evaluate each claim whether it is supported or not. In order to save cost, we randomly sampled 100 samples among the filtered generations. During factuality evaluation, Wikipedia's knowledge source is used in the automatic tool. 

\paragraph{Step 4: Estimation of \textbf{Self-Known} and \textbf{Self-Unknown}} With above decomposed atomic claims, we use GPT-4 Turbo to get question-answer pairs. For each question-answer pair, a prompt template (see \ref{sec:knownUnknown} ) is used to determine whether LLMs consider the proposed answer to be correct. The ratios of supported claims judged as correct, and unsupported claims judged as incorrect are then obtained.

\section{More results}

\begin{table}[h!]
\label{table:llms}
\centering
\scalebox{0.7}{
\begin{tabular}{lcc}
 \multicolumn{1}{l}{}  & \textbf{\#Claims / Gen} &  \textbf{Filtered Rate (\%)} \\ \hline \\
GPT-4 & 60.8 &  12.0 \\
Gemini-1.5-pro &  67.5 &  30.0\\

Claude-3-opus & 41.0 & 42.0 \\ 
Llama-3-70B-Instruct & 45.9 & 17.2 \\ 
Mixtral-8x7b & 44.8 & 0.4 \\ 
Mistral-Large & 48.3 &  5.0 \\ 
\end{tabular}
}
\caption{Statistics for various LLMs when generating
biographical paragraphs. }
\label{tab:llm}
\end{table}
Table~\ref{tab:llm} in the Appendix presents two results for various LLMs: the average number of atomic claims per generation and the filtered rate. The filtered rate represents the percentage of instances where the LLMs do not provide valuable responses, often due to perceiving insufficient information to generate a meaningful answer. We notice that the behavior of Claude-3-opus and Gemini-1.5-pro is more conservative. These models frequently decide not to provide a valuable response, instead stating something like ``I do not have enough verified information''.

\begin{figure*}[t]
\small
\centering
\begin{subfigure}[b]{0.48\textwidth}
   \includegraphics[width=1\linewidth]{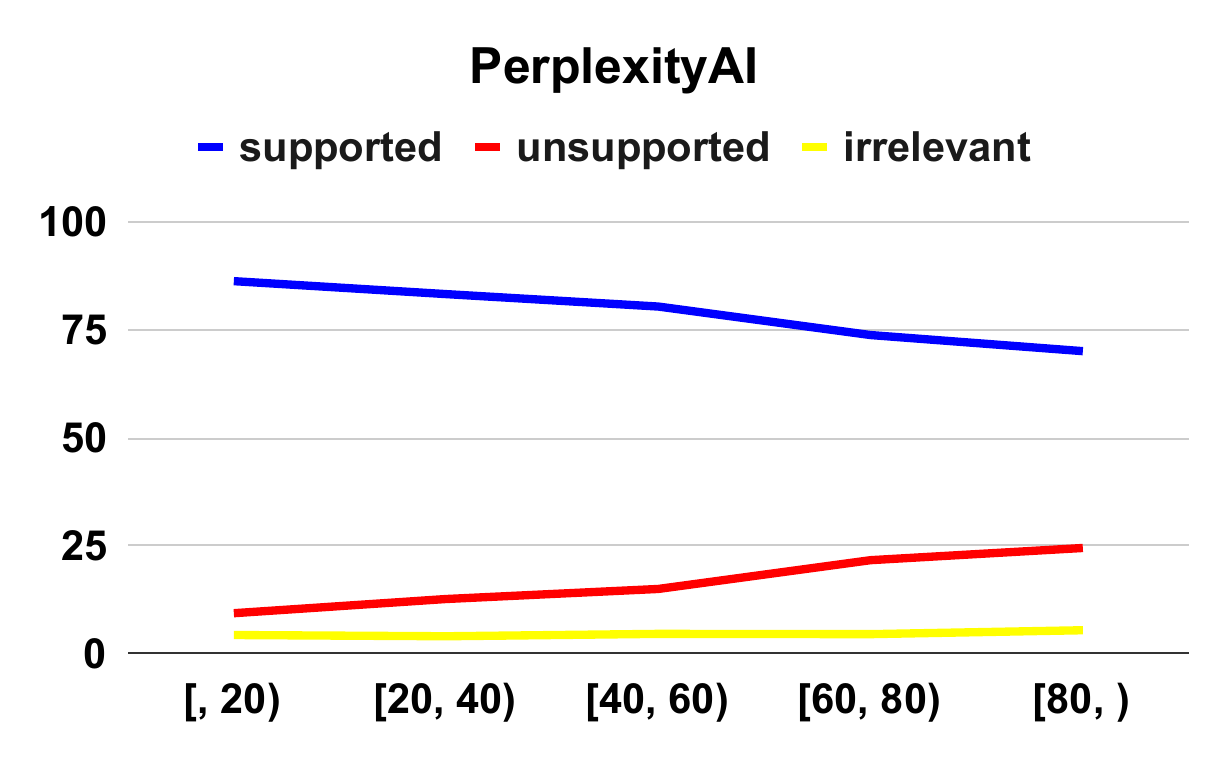}
   \caption{Percentage (\%) of supported, unsupported and irrelevant atomic claims.}
   \label{fig:rNg11} 
\end{subfigure}
\begin{subfigure}[b]{0.48\textwidth}
   \includegraphics[width=1\linewidth]{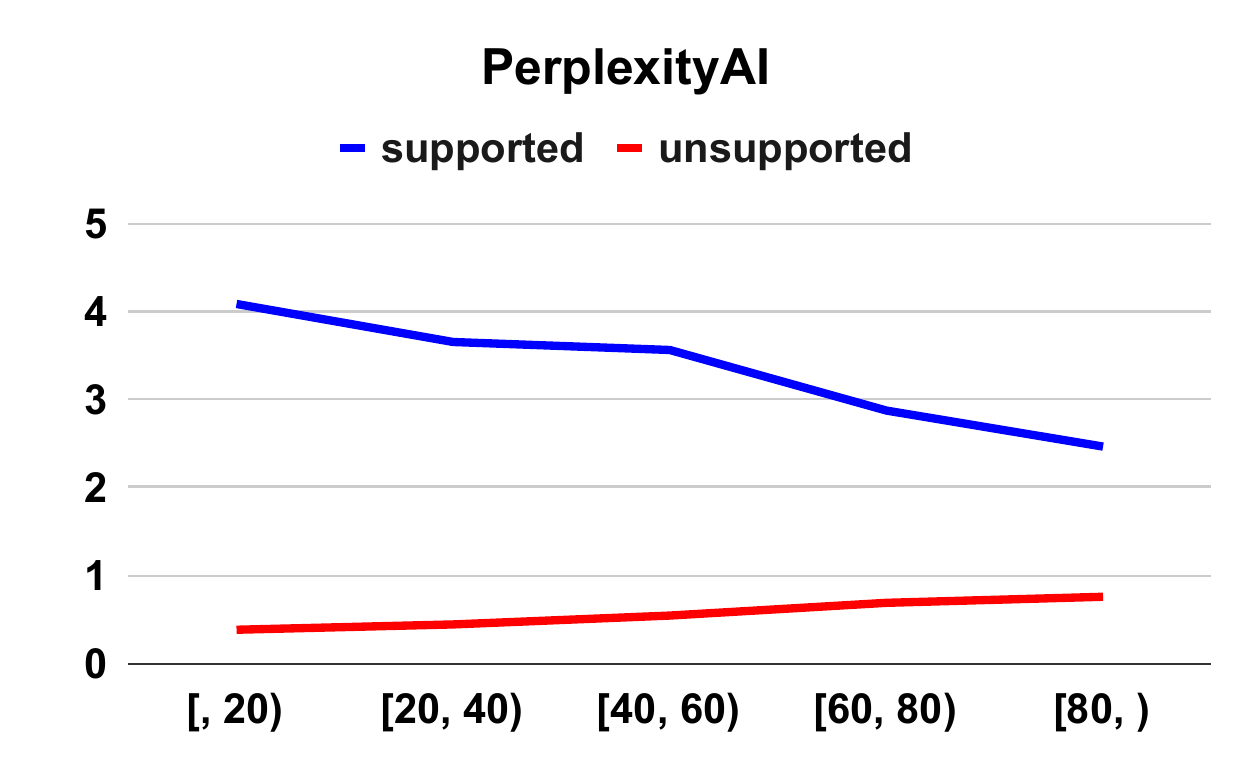}
   \caption{Number of supported and unsupported atomic claims.}
   \label{fig:rNg21}
\end{subfigure}

\caption[Two numerical solutions]{Long-form generation across different relative positions (\%) for PerplexityAI.
}
\label{fig:relative_position_PerplexityAI}
\end{figure*}

\begin{figure*}[h!]
\centering
\centering
  \begin{subfigure}[b]{0.48\linewidth}
    \includegraphics[width=\linewidth]{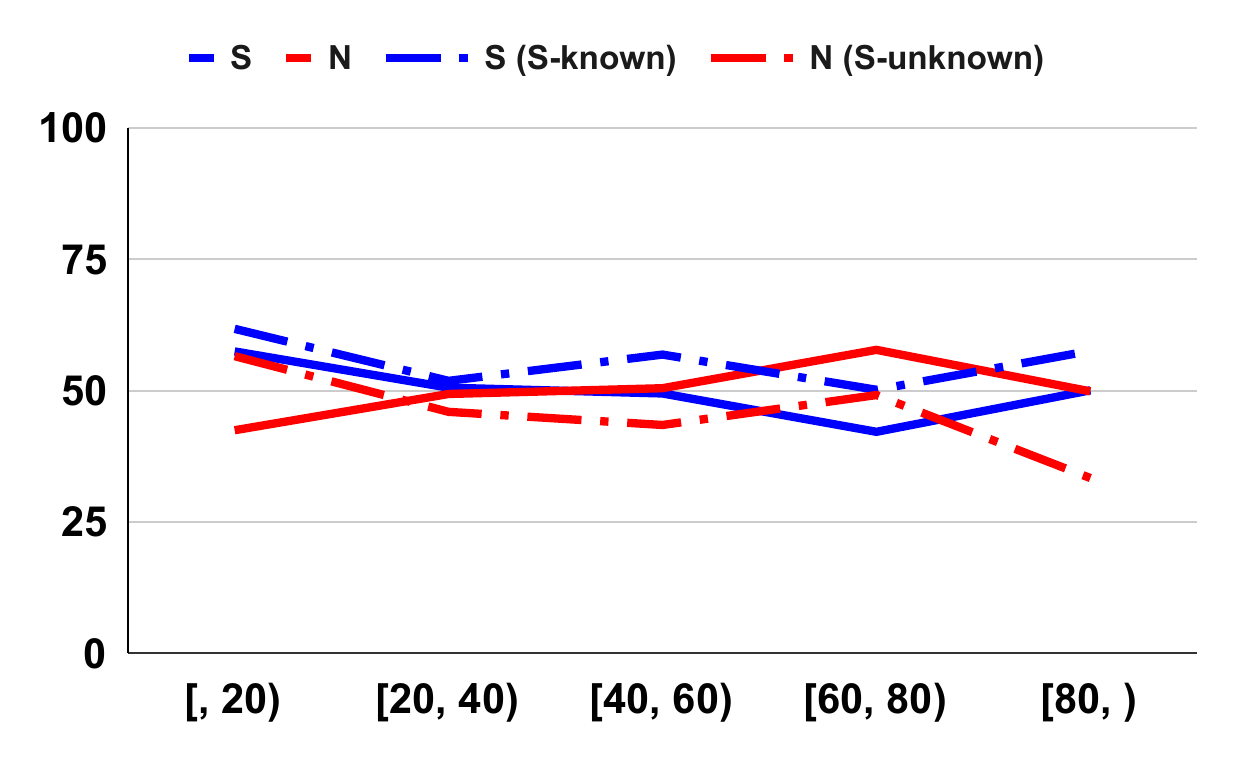}
    \caption{GPT-4}
  \end{subfigure}
    \begin{subfigure}[b]{0.48\linewidth}
    \includegraphics[width=\linewidth]{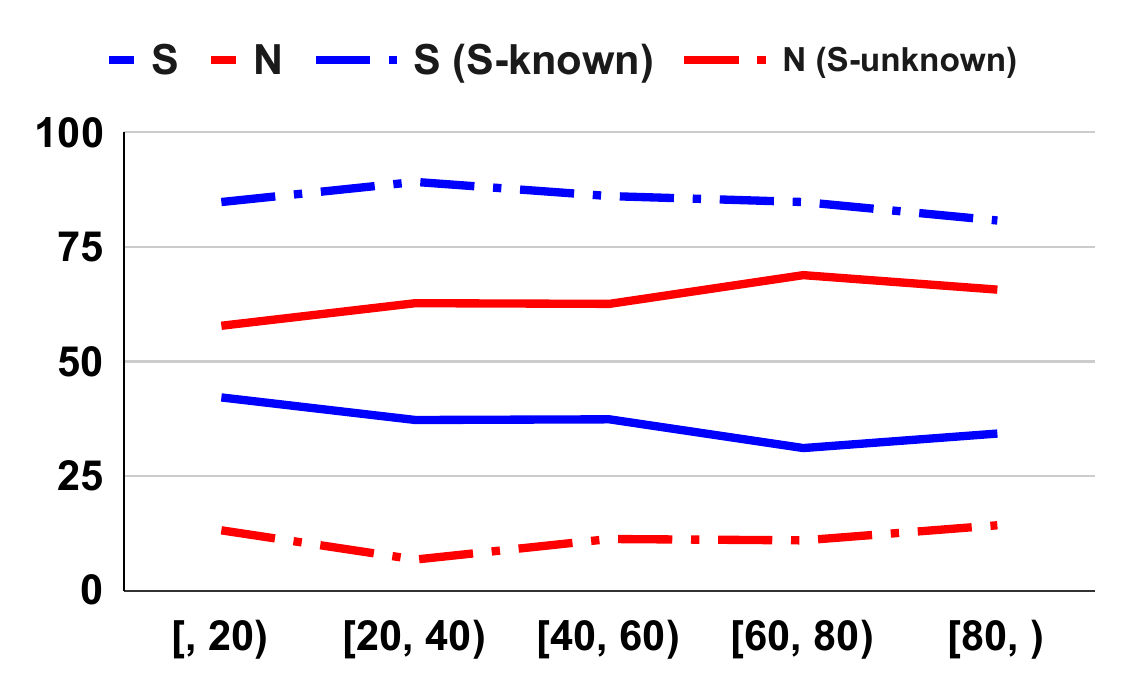}
    \caption{Llama-3-70B-Instruct}
    \end{subfigure}
\caption[Two numerical solutions]{Self-Know and Self-Unknown results of different LLMs across different relative positions (\%). \textbf{S}: \textbf{factuality} (percentage of supported atomic claims); \textbf{N}: percentage of unsupported atomic claims; \textbf{S (S-known)}: percentage of supported atomic claims judged as correct by LLMs; \textbf{N (S-unknown)}: percentage of unsupported atomic claims judged as incorrect by LLMs.}
\label{fig:relative_position_llms_more}
\end{figure*}

\begin{figure*}[h!]
\centering
\centering
  \begin{subfigure}[b]{0.48\linewidth}
    \includegraphics[width=\linewidth]{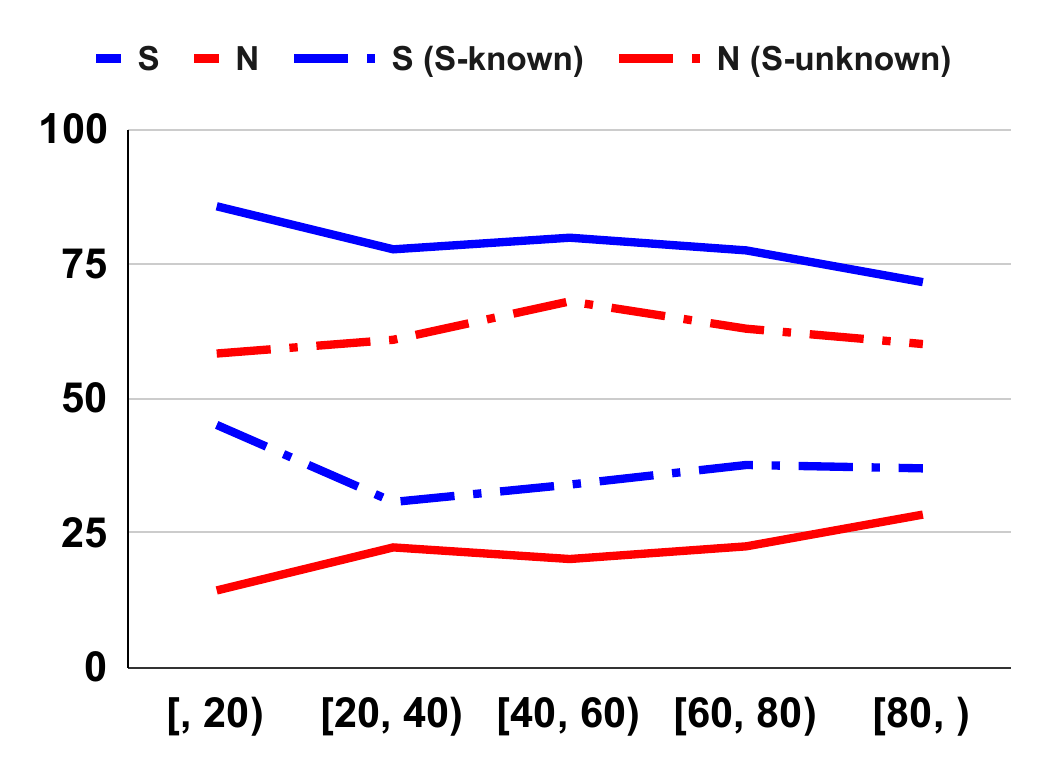}
    \caption{GPT-4-RAG}
  \end{subfigure}
    \begin{subfigure}[b]{0.48\linewidth}
    \includegraphics[width=\linewidth]{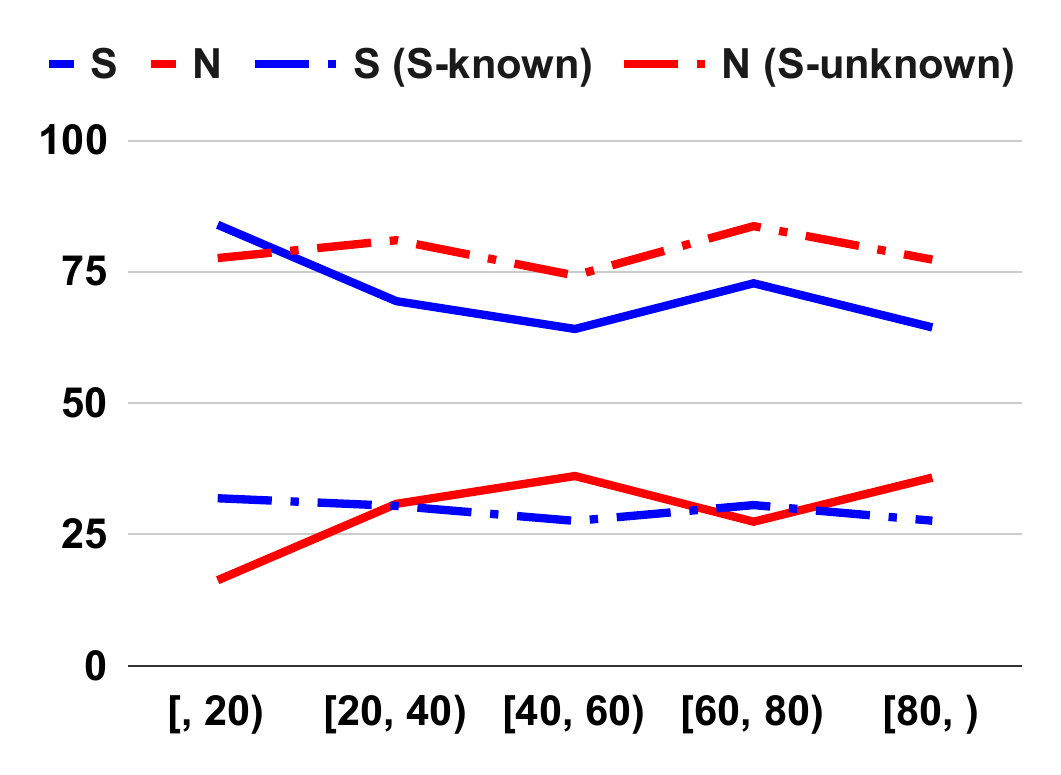}
    \caption{Gemino-1.5-pro-RAG}
  \end{subfigure}
  
\caption[Two numerical solutions]{Self-Know and Self-Unknown results of different \textbf{RAG} models across different relative positions (\%). \textbf{S}: \textbf{factuality} (percentage of supported atomic claims); \textbf{N}: percentage of unsupported atomic claims; \textbf{S (S-known)}: percentage of supported atomic claims judged as correct by LLMs; \textbf{N (S-unknown)}: percentage of unsupported atomic claims judged as incorrect by LLMs.}
\label{fig:relative_position_llms_rag}
\end{figure*}

\begingroup
\begin{table*}[ht]

    \centering
    \begin{tabular}{p{0.96\linewidth}}
    \toprule
    \textbf{Document} [0] Jessie Mae Brown Beavers Jessie Mae Brown Beavers (March 18, 1923 – September 6, 1989) was an American journalist based in Los Angeles, California. She was an editor at the "Los Angeles Sentinel" from 1949 to 1989, and served sixteen years on the city's Human Relations Commission, beginning with her 1973 appointment by mayor Tom Bradley.Early life. Jessie Mae Brown was born in Los Angeles, the daughter of Arnetta Hoyt Brown, a Baptist deaconess. She attended the University of California, Los Angeles, where she earned a bachelor's degree in sociology.Career. Brown was editor of the family section of the "California Eagle" from 1944 to 1949, when she joined the staff of the "Los Angeles Sentinel" as an editor. In 1966 she was one of the organizers and leaders of the Los Angeles chapter of the National Association of Media Women. In 1969, she was given the Outstanding Woman in Journalism Award by the University of Southern California chapter of Theta Sigma Phi. Beavers served on the Los Angeles Human Relations Commission for sixteen years, after she was appointed by mayor Tom Bradley in 1973. On the commission, she worked closely with fellow commissioner Toshiko S. Yoshida during the American bicentennial \\
    \textbf{Document} [1] , chaired the affirmative action subcommittee, and in 1982 organized hearings on racial bias in the entertainment industry. She also served one the Los Angeles County Music and Performing Arts Commission. She ran for a seat on the Los Angeles City Council in 1987. Beavers was a member of Jack and Jill and the Lullaby Guild. In 1968 she was cited for her work by the President's Youth Council. In 1972 she was honored alongside Quincy Jones, editor Ruth Washington, and student leader Willis Edwards by the Committee of Women for Good Government. In 1973 she was honored by colleagues and admirers (including Mayor Bradley, judge Vaino Spencer, columnist Gertrude Gipson, and clubwoman Leontyne Butler King) at a tribute event at the Beverly Wilshire Hotel, hosted by singer Thelma Houston.Personal life. In 1947 Brown married insurance executive Leroy A. Beavers, Jr., nephew of Los Angeles businessman George A. Beavers Jr. Actress Dorothy Dandridge was one of her bridesmaids in the wedding. They had three children. In 1979, she was assaulted and robbed in the driveway of her home. She died in 1989, aged 66 years, in Los Angeles. Her grave is at Forest Lawn Cemetery in Gl\\
    \textbf{Document} [2] endale, California. \\
\midrule
\textbf{Question:} Tell me a bio of Jessie Mae Brown Beavers.\\
\midrule
\textcolor{blue}{Response:} 
Jessie Mae Brown Beavers (March 18, 1923 – September 6, 1989) was a prominent American journalist and community leader based in Los Angeles, California. Born in Los Angeles, she obtained her bachelor's degree in sociology from the University of California, Los Angeles. \\

Beavers's career was deeply intertwined with journalism and \textcolor{red}{activism}. She served as an editor for the \textcolor{yellow}{*California Eagle*} and later the \textcolor{red}{*Los Angeles Sentinel*}, a prominent African American newspaper, for \textcolor{red}{over four decades}. Her commitment to social justice led her to co-found the Los Angeles chapter of the National Association of Media Women and \textcolor{red}{actively advocate for fair representation in the media}. \\

Appointed by Mayor Tom Bradley, Beavers dedicated sixteen years to the Los Angeles Human Relations Commission, tackling issues of racial bias and promoting affirmative action. Her influence extended to her service on the Los Angeles County Music and Performing Arts Commission. \\

Throughout her life, Beavers received numerous accolades, including recognition from the President's Youth Council and a tribute event graced by \textcolor{red}{notable figures like Mayor Bradley} and singer Thelma Houston. \textcolor{red}{Her enduring legacy continues to inspire aspiring journalists and advocates alike.} \\

\bottomrule
\end{tabular}
\caption{Generated outputs of Gemini-1.5-pro: each document represents a retrieval chunk, with text highlighted in red and yellow indicating unsupported claims. Red text represents claims identified as unsupported by the factuality tool. Yellow text represents unsupported claims that are not identified  by the factuality tool.}
\end{table*}
\endgroup
\label{tab:ragExamples}
\end{document}